\documentclass[10pt,twocolumn,twoside]{IEEEtran}
\usepackage{graphicx}
\usepackage{amsfonts,amsmath,amsfonts, amssymb, amstext,bm,amsthm,color}
\usepackage{shortcuts_FW}
\usepackage{booktabs}
\usepackage{float}
\usepackage{cite}
\usepackage{multirow}
\usepackage{algorithm}
\usepackage{algorithmic}
\usepackage[bookmarks=true,colorlinks,linkcolor=black,anchorcolor=black,citecolor=black]{hyperref}  

\newtheorem{ass}{Assumption}

\newtheorem{fact}{Fact}
\newtheorem{remark}{Remark}
\def\mc {\mathcal }
\begin{document}
%
\title{Detection of Insider Attacks in Distributed Projected Subgradient Algorithms}
\author{Sissi~Xiaoxiao~Wu,~Gangqiang~Li,~Shengli~Zhang, and Xiaohui~Lin 
\thanks{This work is supported by the National Natural Science Foundation of China under Grant 61701315; by Shenzhen Technology R$\&$D Fund JCYJ20170817101149906 and JCYJ20190808120415286; by Shenzhen University Launch Fund 2018018.}
\thanks{S. X. Wu, G. Li, S. Zhang and X.~Lin are with the College of Electronics and Information Engineering, Shenzhen University, Shenzhen, China. G. Li is the corresponding author. E-mails: 
{ligangqiang2017@email.szu.edu.cn}, {\{xxwu.eesissi, zsl, xhlin\}@szu.edu.cn}.}
}

%



\maketitle

\begin{abstract}
The gossip-based distributed algorithms are widely used to solve decentralized optimization problems in various multi-agent applications, while they are generally vulnerable to data injection attacks by internal malicious agents as each agent locally estimates its decent direction without an authorized supervision.
In this work, we explore the application of artificial intelligence (AI) technologies to detect internal attacks. We show that a general neural network is particularly suitable for detecting and localizing the malicious agents, as they can effectively explore nonlinear relationship underlying the collected data. Moreover, we propose to adopt one of the state-of-art approaches in federated learning, i.e., a collaborative peer-to-peer machine learning protocol, to facilitate training our neural network models by gossip exchanges. This advanced approach is expected to make our model more robust to challenges with insufficient training data, or mismatched test data. In our simulations, a least-squared problem is considered to verify the feasibility and effectiveness of AI-based methods.
Simulation results demonstrate that the proposed AI-based methods are beneficial to improve performance of detecting and localizing malicious agents over score-based methods, and the peer-to-peer neural network model is indeed robust to target issues.
\end{abstract}

%
\IEEEpeerreviewmaketitle

\section{Introduction}
\IEEEPARstart{r}{ecently}, decentralized optimization algorithm as a popular tool to handle large scale computations has been broadly applied in various fields \cite{nedic2009distributed, nedic2010constrained}. Typical examples of  Internet of Things (IoT) \cite{8721133, 8501581}, multi-agent systems \cite{wu2018review, zhang2018enabling}, wireless communications network \cite{gesbert2007adaptation}, power grid \cite{giannakis2013monitoring}, and federated learning \cite{hegedHus2019gossip}. The design approach in above applications is often refereed as gossip-based optimization problems, wherein interacting agents are randomly selected and exchange information following a point-to-point message passing protocol so as to optimize  shared variables. Aiming at a coordinated response, these agents explicitly disclose their estimates (states) to neighboring agents in each iteration, thereby leading to a consistent globally optimal decision \cite{nedic2009distributed, nedic2010constrained, Tsitsiklis1984}. It is well known that gossip-based algorithms are inherently robust to intermittent communication and built-in fault-tolerance to agent failures. They can also provide a degree of privacy in many applications for participating agents without exchanging local user information \cite{ram2010distributed}. Despite many advantages, these gossip-based algorithms, such as the distributed projected subgradient (DPS) algorithm \cite{nedic2010constrained}, are inherently vulnerable to insider data injection attacks due to the flat architecture, since each agent locally estimates its (sub)gradient without any supervision \cite{sundaram2015consensus, icasspsttackerWu, Li1806:Detecting}. 

Generally speaking, malicious agents' (or attackers) attack on decentralized algorithms depends on specific attacking strategies. Attackers may interfere with distributed algorithms by injecting random data that hinders convergence \cite{yan2012vulnerability}. Especially in an insider attack, the attacker always sends misleading messages to its neighbors to affect the distributed system, resulting in false convergence results \cite{zhao2017resilient, sundaram2018distributed}. For example, a multi-agent system is forced to converge to the target values of attackers in \cite{asilomar} and an average consensus result is disturbed by coordinated attackers in \cite{Gentz2016Data}. 
The attack model we focus on in this work is that the attacker behaves like stubborn agents \cite{mauro03}. To be more specific, they coordinate and send messages to peers that contain a constant bias \cite{sundaram2015consensus, icasspsttackerWu, kailkhura2017data, sundaram2018distributed} and their states can not be changed by other agents.
As studied in \cite{asilomar, Gentz2016Data}, the network always converges to a final state equal to the bias.
This will bring serious security problems to distributed algorithms if the attacker cannot be detected effectively.
Thus, a good defense mechanism is needed to protect these algorithms from internal data injection attacks.

To detect anomalous behaviors in decentralized optimization, one commonly used approach in the literature is to calculate a dependent score through statistical techniques
based on the messages received during the execution of protocol.
For instance, in \cite{yan2012vulnerability}, the authors show that the convergence speed of network will slow down when the attacker is present, and design a score to identify potential attacks.
In \cite{Gentz2016Data}, two score-based detection strategies are proposed to protect the randomized average consensus gossip algorithm from malicious nodes.
In \cite{Kailkhura2015Consensus}, the authors design a comparison score to search for differences between a node and its neighbors, then adjust update rules to mitigate the impact of data falsification attacks.
In \cite{icasspsttackerWu}, the decision score is computed by a temporal difference strategy to detect and isolate attackers.
Similar score-based methods are also shown in \cite{shalom2018detection, ravi2019detection, patel2020distributed}.
While such methods have reasonable performance, the score design is somewhat ad-hoc and relies heavily on the experts to design sophisticated decision functions, and the detection thresholds of these score-based methods need to be adjusted judiciously. 
To circumvent the above difficulties, our idea in this work is to utilize the artificial intelligent (AI) technology to approximate more sophisticated decision functions.
It is worth mentioning that AI technology has succeed in many applications with the same purpose, including image recognition \cite{he2016deep}, natural language processing \cite{devlin2018bert}, power grid \cite{wang2018deep}, and communications \cite{yu2019deep}.  
Furthermore, AI also plays an important role in network security \cite{wu2019survey}, such as anomaly intrusion detection \cite{Doboli2007Discovery}, malicious PowerShell \cite{rusak2018ast}, distributed denial of service (DDoS) attacks \cite{rahman2019ddos} and malicious nodes in communication networks \cite{li2018detecting, li2020neural}. 

The main purpose of this work is to apply AI technology to address the problem of detecting and localizing attackers in decentralized gossip based optimization algorithms. 
While our AI-based methods and training philosophy can be applied to a wide set of multi-agent algorithms and attack scenarios, we focus on testing the approach on a case that has been thoroughly studied in \cite{icasspsttackerWu,li2020detect}, to facilitate the comparison.
Concretely, we proposed two AI-based strategies, namely the temporal difference strategy via neural networks (TDNN) and the spatial difference strategy via neural networks (SDNN). We will show that even basic neural network  (NN) models exhibit a good ability to extract non-linearity from our training data and thus can well detect and localize attackers, given that 1) the collected training data can well represent the attack model, and 2) training data from all agents can be fully learned at the training center.

Unfortunately, collecting good and enough data which perfectly fits the real attack model is usually difficult. First of all, due to the intrinsic of gossp-algorithm, it is difficult and expensive to collect sufficient training samples at each agent.  Also, with the emergence of new large-scale distributed agents in the network, sometimes it is hard to upload decentralized data at each agent to a fusion center due to storage and bandwidth limitations~\cite{giaretta2019gossip}. Furthermore, as the insider attacks could occur at any agent in the network, the training data may not cover all the occurrences of the attack. Therefore, some individually trained NN model at each agent may not fit in all insider attack events.
A new approach to alleviate these issues is to leverage the decentralized federated learning \cite{savazzi2020federated}, which utilizes the collaboration of agents to perform data operations inside the network by iterating local computations and mutual interactions. 
Such a learning architecture can be extremely useful for learning agents with access to only local/private data in a communication constrained environment \cite{jiang2017collaborative}.
Specially, as one of the state-of-the-art approach in decentralized federated learning, gossip learning is very suitable for training NN models from decentralized data sources \cite{ormandi2013gossip}, with the advantages of high scalability and privacy preservation. 
Thus, we propose a collaborative peer-to-peer learning protocol to help training our NN models by gossip exchanges.
Specifically, each agent in the network has a local model with the same architecture, and only relies on local collaboration with neighbors to learn model parameters. 
It is worth noting that in this process each agent trains the local model by its local data periodically, and then send the local model parameters to its neighbors. 
It is expected that each agent can learn a local model close to the \emph{global model} (i.e., a NN trained by the center, which contains training data from all agents), so as to provide robustness in the case of insufficient and mismatched local data.

It is also worth mentioning differences between this work and some previous work. Previous work \cite{Gentz2016Data} aims at the score-based method for securing the gossip-based average consensus algorithm. \cite{li2020neural} improves the score-based method by using AI-based methods
while it still targeted at an average consensus algorithm. We remark that The inputs for AI model in \cite{li2020neural} does not always work for optimization algorithms. \cite{icasspsttackerWu,li2020detect} provide some preliminary results for protecting optimization algorithms while it only focus on partial neighboring information. This work is the first one which well elaborates AI-based methods for a DPS algorithm using full information from neighboring signal. More importantly, the proposed collaborative learning method is novel and effective to make the defense model more robust to different events of attacks, making our models more practical to multi-agent applications. In summary, the proposed AI-based strategies have following characteristics: 1) they can automatically learn appropriate decision models from the training data, thus reducing the dependence on complicated pre-designed models;
2) they adaptively scale the decision thresholds between 0 and 1, which reduces the difficulty of threshold setting;
3) they improve the performance of detecting and localizing attackers and show good adaptability to different degree of agents.
4) they have strong robustness to the scenarios with insufficient training data, or mismatched training data. Preliminary numerical results demonstrate that the proposed AI-based strategies are conducive to solve the insider attack problem faced by the DPS algorithm.


The rest of the paper is organized as follows. In Section \ref{sec:system}, we describe the decentralized multi-agent system and the attack scheme against the DPS algorithm.
In Section \ref{sec:strategies}, we review score-based strategies and propose two AI-based defense strategies to detect and locate attackers.
Section \ref{sec:manner} introduces a collaborative peer-to-peer training protocol for NN, dealing with insufficient samples or mismatched samples available on different agents. 
Simulation results are given in Section \ref{sec:results} to confirm the effectiveness of the proposed strategies.
We conclude this work in section \ref{sec:conclusion}.
\section{System Model} \label{sec:system}
We consider a multi-agent network which can be defined by an undirected graph $\mathcal G=(\mathcal V,\mathcal E)$, wherein $\mc V=\{1,\cdots,n\}$ represents the set of all agents and $\mc E \subseteq \mc V \times \mc V$ represents the set of all edges. We define the set of the neighbor nodes of an agent $i\in \mc V$ by $\mc N_i=\{v_j\in \mc V: (v_i, v_j) \in \mc E\}$. All the agents in the distributed network follow a gossip-based optimization protocol; see Algorithm \ref{alg:1}. That is, in each iteration of information exchange, an agent only directly communicates with its neighbors. We thus define a time-varying graph as ${\mathcal G}(t) \eqdef (\mathcal V,\mathcal E(t))$ for the $t$th iteration
and the associated weighted adjacency matrix is denoted by ${\bm A}(t) \in \RR^{n \times n}$, where $[ {\bm A}(t) ]_{ij} \eqdef A_{ij}(t) = 0$ if $(v_j,v_i) \notin \mathcal E(t)$. 
For this network with $n$ agents, we have the following assumption:
\begin{ass} \label{ass:graph}
 { There exists a scalar $\zeta \in (0,1)$ such that for all $t \geq 1$ and $i=1,\cdots,n$ :} 
	\begin{itemize}
	\item $A_{ij}(t) \geq \zeta$ if $(i,j) \in \mathcal E(t)$, 
	\item $\sum_{i=1}^{n}A_{ij}(t)=1$, $A_{ij}(t)=A_{ji}(t)$;
	\item The graph $(\mathcal V, \cup_{\ell=1}^{B_0} \mathcal E(t+\ell))$ is connected for $B_0 < \infty$.
	\end{itemize}
\end{ass}

The goal of these agents is cooperatively to solve the following optimization problem:
\begin{equation} \label{eq:opt}
\min_{\bm x} f(\bm x) \eqdef \frac{1}{n} \sum_{i=1}^n f_i ( \bm x )~~\text{s.t.}~~\bm x \in X \eqs.
\end{equation}
where $X \subseteq \RR^d$ is a closed convex set common to all agents and $f_i: \RR^d \rightarrow \RR$ is a local objective function of agent $i$. Herein, $f_i$ is convex and not necessarily differentiable function which is only known to agent $i$. 

In this setting, we denote the optimal value of problem \eqref{eq:opt} by $f^\star$. A decentralized solution to estimate $f^\star$ of this problem is the DPS algorithm \cite{nedic2010constrained}. In this algorithm, each agent locally updates the decide variable by fusing the estimates from its neighbors and then take the subgradient of this function at the updated decide variable to be the decent direction for the current iteration. To be more specific, when applied this algorithm to solve problem \eqref{eq:opt}, it performs the following iterations:
\begin{equation}\label{eq:dgd}
\begin{split}
& \bar{\bm x}_i(t) = \sum_{j=1}^n A_{ij}(t) \bm x_j(t) \eqs.\\[.1cm]
& \bm x_i(t+1) = {P}_{X} \big[ \bar{\bm x}_i(t)  - \gamma(t) {\hat \grd} f_i \big( \bar{\bm x}_i(t) \big) \big] \eqs. 
\end{split}
\end{equation}
for $t \geq 1$, where $A_{ij}(t)$ is a non-negative weight and $\gamma(t) > 0 $ is a diminishing stepsize.
${P}_{X} \big[ \cdot \big]$ denotes the projection operation onto the set $X$ and $\hat \grd f_i \left( \bar{\bm x}_i(t) \right)$ is a subgradient at agent $i$ of the local function $f_i$ at $\bm x= \bar{\bm x}_i(t)$. Then, we have the following result:
\begin{fact} \label{fact:dgd}
\cite{nedic2010constrained} Under Assumption \ref{ass:graph}. 
If $\| \hat \grd f_i ( \bm x) \| \leq C_1$ for some $C_1$ and for all $\bm x \in X$, and the step size satisfies $\sum_{t=1}^\infty \gamma(t)
= \infty$, $\sum_{t=1}^\infty \gamma^2(t) < \infty$, then for all $i,j \in \mathcal V$ we have
\beq
\lim_{t \rightarrow \infty} f( \bm x_i (t) ) = f^\star~~\text{and}~~\lim_{t \rightarrow \infty}
\| \bm x_i (t) - \bm x_j(t) \| = 0 \eqs \notag.
\eeq
\end{fact}
The above fact tells that for these convex problems,  the DPS method will converge to an optimal solution of problem \eqref{fact:dgd}. In the next, we will discuss how this convergence will change when there is attack within the network.

\begin{algorithm}[t!] 
\caption{The gossip-based optimization protocol}\label{alg:1} 
\par\setlength\parindent{1em}\textbf{Input}: ~~Number of instances $K$, and iterations $T$.

\begin{algorithmic}
\FOR{$k=1,\cdots,K$}
		\item Initial states: $\bm x_i^k(0)=\bm \beta_i^k~\forall~i \in {\mc V}$
		\FOR{$t=1,\cdots,T$}
				 \item $\bullet$ Uniformly wake up a random agent $i\in {\mc V}$
				 \item $\bullet$ Agent $i$ selects agent $j \in {\mc N}_i $ with probability $P_{ij}$
				 \item $\bullet$ The trustworthy agents $i,j \in \mc V_t$ update the states \par\setlength\parindent{1em} according to the rules in \eqref{eq:dgd}.
				 \item $\bullet$ The malicious agents follow the attack scheme to \par\setlength\parindent{1em} keep their original states, as seen in \eqref{eq:attacker}.
		\ENDFOR 
\ENDFOR 
\end{algorithmic} 
\end{algorithm}
\subsection{Data Injection Attack From Insider}
In this setting, we assume that the set of agents $\mc V$ can be divided into two subsets: the set of trustworthy agents ${\mathcal V_t}$ and the set of malicious agents (attackers) ${\mathcal V_m}$, as seen in Fig. \ref{fig:task}. We have ${\mathcal V} = {\mathcal V}_t \cup {\mathcal V}_m$ and $n=|{\mathcal V}_t|+|{\mathcal V}_m|$. 
In our attack model, attackers are defined as agents whose estimates (or states) can not be affected by other agents, and those coordinated attackers try to drag the trustworthy agents to their desired value. 
If $i \in {\mathcal V}_t$, a trustworthy agent  will perform the rules in \eqref{eq:dgd}. 
Otherwise, an attacker $j \in {\mathcal V}_m$ will update its state with the following rule:  
\beq \label{eq:attacker}
    \bm x_j(t) = {\bm \alpha} + {\bm r}_j(t),~\forall~j \in \mathcal V_m \eqs.
\eeq
where $\bm {\alpha}$ is the target value of attackers, ${\bm r}_j(t)$ is an artificial noise generated by attackers to confuse the trustworthy agents. If there are more than one attacker in the network, we assume that they will coordinate with each other to converge to the desired value ${\bm \alpha}$.  Meanwhile, to disguise attacks, they will independently degenerate artificial noise ${\bm r}_j(t)$ which decays exponentially with time,
i.e., $ \lim_{t \rightarrow \infty} \| {\bm r}_j (t ) \| = 0$  for all $j\in \mathcal V_m$. 

For the time varying network, let $\mathcal E(\mathcal V_t;t)$ be the edge set of the subgraph of $\mathcal G(t)$ with only the trustworthy agents in $\mathcal V_t$. The following assumption is needed to ensure a successful attack on the DPS algorithm:

\begin{ass} \label{ass:att}
There exists $B_1, B_2 < \infty$ such that for all $t \geq 1$, $1$) the composite sub-graph $(\mathcal V_t, \cup_{\ell=t+1}^{t+B_1} \mathcal E(\mathcal V_t;\ell))$ is connected; $2$) there exists a pair $i \in \mathcal V_t$, $j \in \mathcal V_m$ with $(i,j) \in \mathcal E(t) \cup \ldots \cup \mathcal E(t+B_2-1)$. 
\end{ass}
Based on this assumption, we have the following fact:
\begin{fact} \label{fact:att}
\cite{icasspsttackerWu} Under Assumptions~\ref{ass:graph} and \ref{ass:att}.
If $\| \hat \grd f_i ( \bm x) \| \leq C_2$ for some $C_2$ and for all $\bm x \in X$, and $\gamma(t) \rightarrow 0$, we have:
\beq \label{eq:converge}
\lim_{t \rightarrow \infty} \max_{i \in \mathcal V_t} \| \bm x_i (t) - {\bm \alpha} \| = 0 \eqs.\notag
\eeq 
\end{fact}
This fact implies that in this attack scheme, the attackers will succeed in steering the final states.
This was also proved in our previous work \cite{icasspsttackerWu}.

\section{Detection and Localization Strategies} \label{sec:strategies}
\begin{figure}[t]
\begin{minipage}[b]{1.0\linewidth}
  \centering
	  \centerline{\includegraphics[width=8cm]{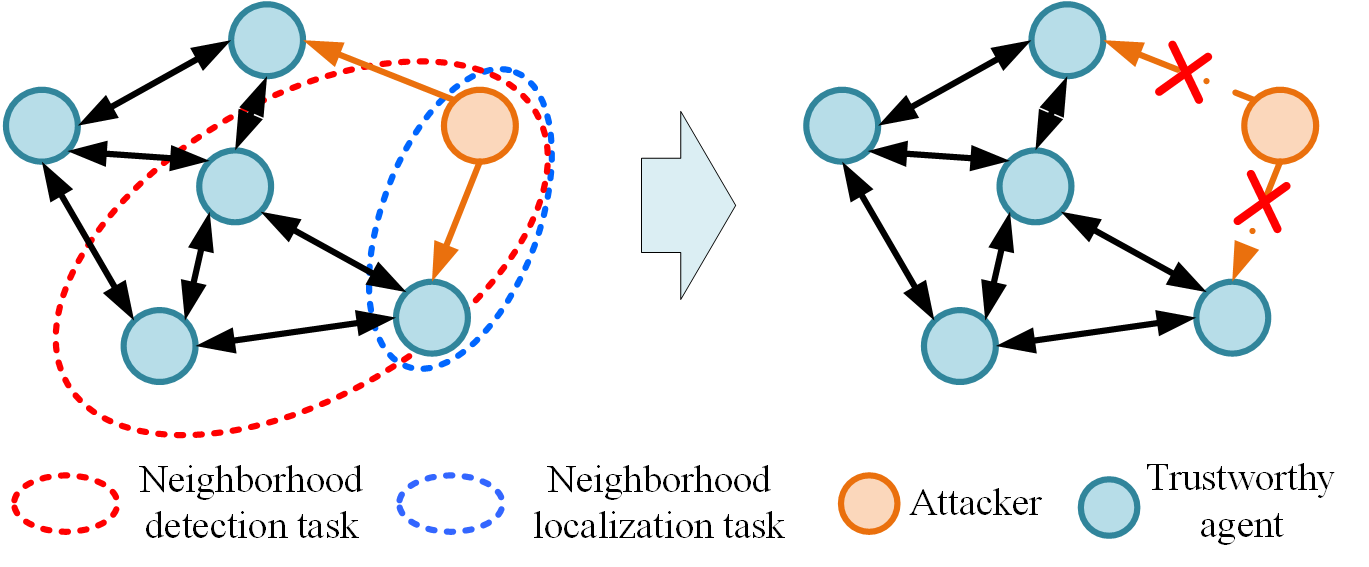}}
\end{minipage}
\caption{Neighborhood tasks in the attack detection scheme. Each trustworthy agent performs ND and NL tasks independently for isolating attacker from the network.}
\label{fig:task}
\end{figure}

The DPS algorithm runs in a fully decentralized fashion in trustworthy agents $i \in \mc V_t$. The {neighborhood detection} (ND) task and {neighborhood localization} (NL) task are then introduced for detecting and localizing attacker. To facilitate our discuss, we consider the following hypotheses. The ND task is defined as follows:
\begin{equation} \label{ND_task}
  \begin{split}
        {\cal H}_0^{i}:&~ {\mathcal N}_i \cap {\mc V}_m = \emptyset,    \quad \text{No neighbor is an attacker},\\
        {\cal H}_1^{i}:&~ {\mathcal N}_i \cap {\mc V}_m \neq \emptyset, \quad \text{At least one neighbor is the attacker},
  \end{split}
\end{equation}
where ${\cal H}_0^{i}$ and ${\cal H}_1^{i}$ as two events of agent $i$ for the ND task. When event ${\cal H}_1^{i}$ is true at agent $i$, the second task is to check if the neighbor $j \in \mc N_i$ is an attacker. The NL task is defined as follows:
\begin{equation} \label{NL_task}
  \begin{split}
	      {\cal H}_0^{ij}:&~ j \notin {\mathcal V}_m, \quad \text{Neighbor $j$ is not an attacker}, \\
        {\cal H}_1^{ij}:&~ j \in {\mathcal V}_m, \quad \text{Neighbor $j$ is an attacker},~~~~          
  \end{split}
\end{equation}
where ${\cal H}_0^{ij}$ and ${\cal H}_1^{ij}$ are as two events of agent $i$ for the NL task. If event ${\cal H}_1^{ij}$ is true, we say that the attacker is localized. We remark that such hypotheses were also made in previous work \cite{Gentz2016Data, li2020detect}.
An illustration of the neighborhood detection and localization tasks is shown in Fig. \ref{fig:task}. Notice that the NL task is executed only if the event ${\cal H}_1^{i}$ in the ND task is true. Moreover, once the attacker is localized, trustworthy agents will disconnect from the attacker in the next communication. In this way, it is expected that the network can exclude all the attackers from the network.

To proceed our tasks, we run the asynchronous gossip-based optimization algorithm (Algorithm \ref{alg:1}) for $K$ instances.
We denote $\tilde{\bm X}_{i}^k$ as the neighborhood state matrix collected by agent $i$ in $k$th instance, i.e., $k \in [1,\cdots,K]$. 
The ND and NL tasks can be described as follows:
\begin{align}
      \tilde{\bm X}_{i}^k \eqdef & [\bm x_i^k, \bm x_1^k,  \cdots,\bm x_j^k,\cdots,\bm x_{\left|\mathcal N_i\right|}^k]^\top~\forall~ j \in {\cal N}_i, \\
     & {y}_i = {\mathrm {F_{ND}}}(\tilde{\bm X}_{i}^1,\cdots,\tilde{\bm X}_{i}^K) \overset{{\cal H}_1^{i}}{\underset{{\cal H}_0^{i}}{\gtrless}} \delta,\\
		 &	z_{ij} = {\mathrm {F_{NL}}}(\tilde{\bm X}_{i}^1,\cdots,\tilde{\bm X}_{i}^K) \overset{{\cal H}_1^{ij}}{\underset{{\cal H}_0^{ij}}{\gtrless}} \epsilon.
\end{align}
where $\bm x_j^k \in \RR^d$ is the state vector of agent $j \in \mc N_i$, which can be directly obtained by agent $i \in \mc V_t$ from its neighbors,
$y_i \in \RR$ is a metric that indicates whether an attacker is present in the neighborhood of agent $i$, and $\bm z_{ij}=[z_{i1},\cdots,{z}_{i\left|\mathcal N_i\right|}]^\top \in \RR^{\left|\mathcal N_i\right|}$ is the metric vector for localization task. Herein,
$\delta > 0$ and $\epsilon > 0$ are some pre-designed thresholds. 

On top of the detection and localization strategies, we have an important assumption about the initial states:
\begin{ass} \label{ass:3}
We have the prior information about the expected initial states about the mean of attackers $\EE [\bm x_{j}^{k}(0)] = \bar{\bm \alpha}, j\in \mc V_m$ and trustworthy agents $\EE [\bm x_{i}^{k}(0)]=\bar{\bm \beta}, i\in \mc V_t$ .  Moreover, $\bar{\bm \alpha} \neq \bar{\bm \beta}$ in general.
\end{ass}
Note that this assumption is practical as the attacker always aims at dragging the trustworthy agents to its desired value, which is usually different from the optimal solution. Otherwise, we may not consider it as a meaningful attack.

\begin{remark}
    ${\mathrm {F_{ND}}}(\cdot)$ and ${\mathrm {F_{NL}}}(\cdot)$ are statistical decision functions judiciously designed for ND and NL tasks respectively.  
     For each agent $i \in \mc V_t$, these decision functions are used to calculate the criterion metrics to identify attackers. 
\end{remark}
\subsection{The Score-based Method}
As a remedy to protect these distributed optimization algorithms, such score-based methods have been studied in \cite{asilomar, Gentz2016Data}, which stem from statistical techniques. 
For gossip-based DPS algorithm, a temporal difference strategy (TD) in \cite{icasspsttackerWu} and a spatial difference strategy (SD) in \cite{li2020detect} are proposed to detect and localize the attackers, and these two strategies are reviewed below. 

\subsubsection{Temporal Difference Strategy} Since the expected initial states about the mean of attackers and trustworthy agents are different,
when $t \rightarrow \infty$, the network will be mislead by the attackers to $\EE [\bm x_{j}^{k}(\infty)] = \bar{\alpha} =\EE [\bm x_{i}^{k}(\infty)]$. 
This implies that the difference between the initial state and the steady state can be used to detect anomalies. For each trustworthy agent $i \in \mc V_t$, the following score can be evaluated \footnote{For each instance $k$, each agent evaluates $\Delta_j(t) \triangleq \bm x^{k}_j(t) - \bm x_j^k(t-1)$ at iteration $t$ and sums it over all the iterations to obtain $\big(\bm x^{k}_j(T) - \bm x_j^k(0)\big)$}:
\beq \label{eq:tdm} 
    {\xi}_{ij} \eqdef \frac {1} {Kd} \sum_{k=1}^K {\bm 1^{\top}\big(\bm x^{k}_j(T) - \bm x_j^k(0)\big)}, j \in {\cal N}_i.
\eeq
Herein, $T \rightarrow \infty$ is sufficiently large, $d$ is the state dimension of agents, $\bm 1$ is an all-one vector. $\bm x^{k}_j(T)$ and $\bm x_j^k(0)$ are respectively the last and the first state for agent $j$ observed by agent $i$. To discern the events in ND task, the detection criterion is defined as follow: 
\beq \label{eq:tdm1} 
    \hat{y}_i  \eqdef \frac{1}{|{\mathcal N}_i|} \sum_{j \in {\mathcal N}_i} \left| \big( \xi_{ij} -  \overline{\xi}_i \big) \right| \overset{{\mathcal H}_0^i}{\underset{{\mathcal H}_1^i}{\lessgtr}} \delta_{\mathrm {TD}}.
\eeq
where $ \overline{\xi}_i=1/\left|\mc N_i \right| \sum_{j\in \mc N_i } \xi_{ij}$ is the average of neighborhood of agent $i$.
Intuitively, $\EE[ \hat{y}_i ] = 0$ when the event $\mc H_0^{i}$ is true, otherwise $\EE[ \hat{y}_i ] \neq 0$ when the event $\mc H_1^{i}$ is true.
$\delta_{\mathrm {TD}}$ is a pre-designed threshold of the ND task.

For the NL task, these two events $\mc H_1^{ij}$ and $\mc H_0^{ij}$ are checked by the following criterion:
\beq\label{eq:tdm2}
    \hat{z}_{ij} \eqdef |{\xi}_{ij}| \overset{{\cal H}_1^{ij}}{\underset{{\cal H}_0^{ij}}{\lessgtr}} \epsilon_{\mathrm {TD}},~\forall~ j \in {\cal N}_i \eqs.
\eeq
Herein, $\epsilon_{\mathrm {TD}}$ is a pre-designed threshold used to identify which neighbor is the attacker. Note that $\EE[\hat{z}_{ij}]$ is close to 0 if an agent $j$ is an attacker, seen in \eqref{eq:tdm}.

\subsubsection{Spatial Difference Strategy}
According to \eqref{eq:attacker}, attackers always try to mislead the network to their desired value, and thus the transient state in the network will also be affected during the attack process. 
Unlike the TD method that only uses the initial state and steady state, the transient states are considered in the SD method for better performance.
We expect that the expected state $\EE[\bm x_i^k(t)-\bm x_j^k(t)]$ between neighbor $j$ and monitoring agent $i$ will behave differently in events $\mc H_0$ and $\mc H_1$, i.e., $j\in \mc N_i$ and $0<t<\infty$. For the ND task, agent $i$ evaluates the following metrics:
\begin{equation}\label{eq:sdm}
     \overline{\bm \varphi}_{ij}^k \eqdef \sum_{t=0}^{T} \Big(\bm x_j^k(t) - \overline{\bm x}_i^k(t) \Big), j\in \mc N_i.
\end{equation}
\beq \label{eq:sdm1}
\check{y}_i  \eqdef \frac{1}{|{\cal N}_i|} \sum_{j \in {\cal N}_i}  \Big( \frac{1}{Kd} \sum_{k=1}^K {\bm 1^{\top} \overline{\bm \varphi}_{ij}^k} \Big)^2 \overset{{\cal H}_0^{i}}{\underset{{\cal H}_1^{i}}{\lessgtr}} \delta_{\mathrm {SD}}.
\eeq
where
\beq \label{eq:avg}
\overline{\bm x}_i^k(t)=1/\left|\mc N_i \cup i \right| \sum_{j\in \{\mc N_i \cup i\}} \bm x_j^k(t)
\eeq 
is the neighborhood average of agent $i$ when iterating $t$ in instance $k$. 
$\overline{\bm \varphi}_{ij}^k$ is the sum of differences between neighbor agent $j$ and the neighborhood average $\overline{\bm x}_i^k(t)$ in all iterations.
$\delta_{\mathrm {SD}}$ is a pre-designed threshold.

For the NL task, we compare the state of neighbor agent $j$ with agent $i$ to check the events in \eqref{NL_task}.
The following criteria are used:
\beq 
\bm \varphi_{ij}^k \eqdef  \sum_{t=0}^{T} \big( \bm x_j^k(t) - \bm x_i^k(t) \big) - \overline{\bm \varphi}_{ii}^k, j\in \mc N_i.
\eeq
\beq\label{eq:sdm2} 
\check{z}_{ij} \eqdef \Big( \frac{1}{Kd} \sum_{k=1}^K {\bm 1^{\top}   \bm\varphi_{ij}^k} \Big)^2 \overset{{\cal H}_0^{ij}}{\underset{{\cal H}_1^{ij}}{\lessgtr}} \epsilon_{\mathrm {SD}}, \forall j\in \mc N_i.
\eeq
where $\bar{\bm \varphi}_{ii}^k$ is calculated by agent $i$ itself, seen in \eqref{eq:sdm}. $\bm \varphi_{ij}^k$ is the metric between agent $i$ and agent $j$, and $\epsilon_{\mathrm {SD}}$ is a pre-designed threshold used to identify the attacker.
\subsection{The AI-based Method}\label{sec:AI_methods}
\begin{figure}[t]
\begin{minipage}[b]{1.0\linewidth}
  \centering
	  \centerline{\includegraphics[width=8cm]{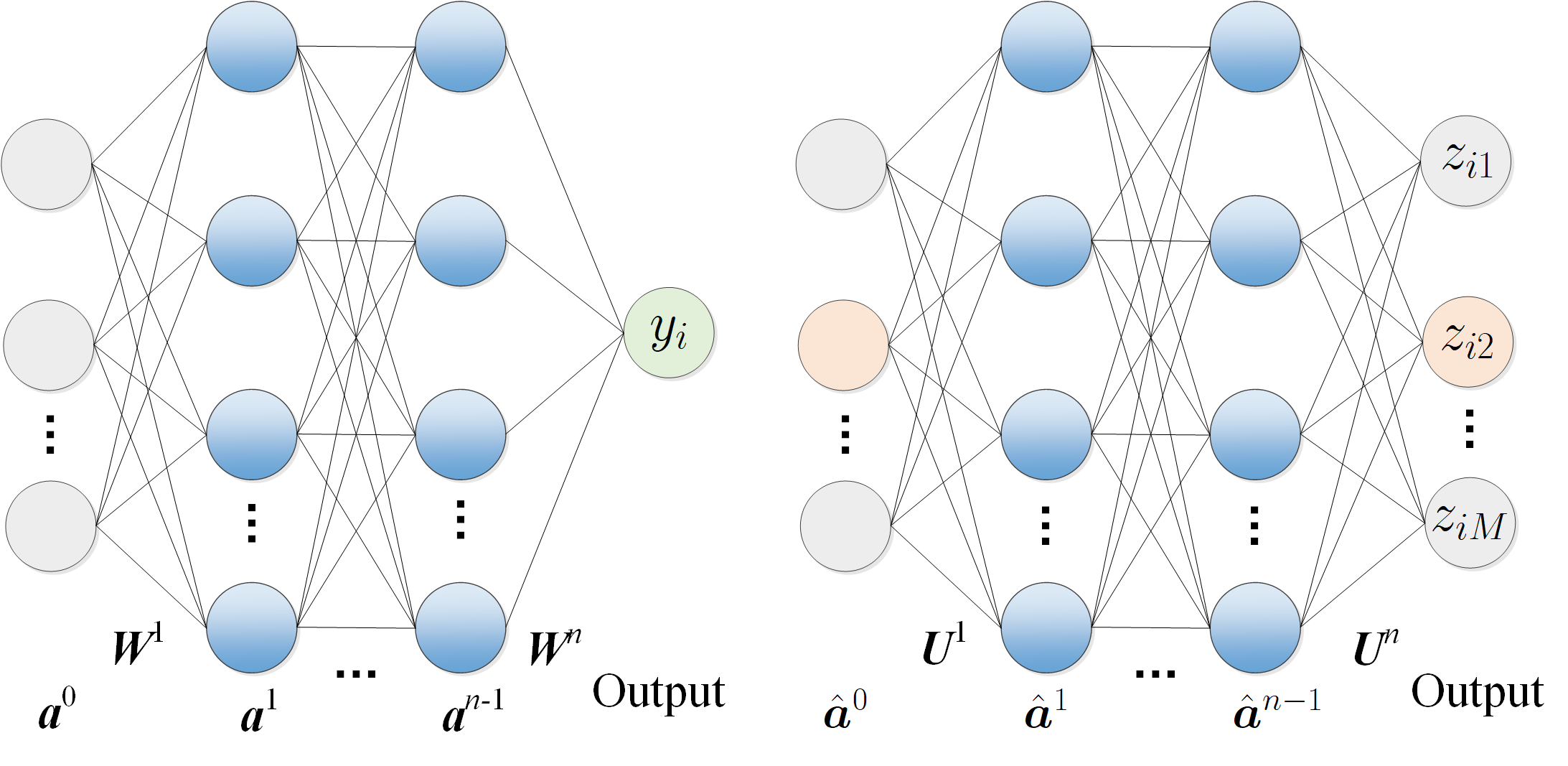}}
\end{minipage}
\caption{The TDNN method at trustworthy agent $i$: (Left) NN for ND task, (Right) NN for NL task. SDNN shares a similarly structure with TDNN.}
\label{fig:nn_network}
\end{figure}
In fact, the reason why \eqref{eq:tdm1}, \eqref{eq:tdm2}, \eqref{eq:sdm1} and \eqref{eq:sdm2} take effect is that the anomalies will cause the the measured metrics to behave statistically different.
In such score-based methods, these decision functions of ND and NL tasks are approximately linear or quadratic functions, which fuse the states obtained by agent $i$ into a scalar score for classification.
A natural question that follows is whether there exists more sophisticated nonlinear functions that can better classify those events in the two neighborhood tasks.
This is a natural application of AI technology for learning the complex mapping relationships in a classification problem.

In the following, we propose to apply NN to handle the ND and NL tasks. Let $M={\mathrm {max}}_i \left|\mc N_i\right|$ be the input dimension of these NNs. Then, NNs can be trained at each monitoring agent in an offline manner using data collected from each agent.
To facilitate our approach, we use the following process to collect training data for the AI-based methods:
\begin{ass} \label{ass:net}
	Assume that we have set a training data collecting process which contains $P$ training network $\mathcal G_p=(\mathcal V_p,\mathcal E)$ for $p={1, 2,.., P}$. For each network $\mathcal G_p$, a randomly chosen agent\footnote{There could be more than one attackers in the training network while herein we only consider the simplest case.} takes the role as an attacker. Based on Assumption \ref{ass:3}, we run the asynchronous gossip-based optimization algorithm (Algorithm \ref{alg:1}) for $K$ instances and record $\tilde{\bm X}_{i}^k$ as the data samples with the ground truth label  `1' for event $\mathcal H_1^i$ where agent $i$ is either in the neighborhood of (next to) the attacker or beyond the neighborhood of (far from) the attacker. 
\end{ass}
We remark here that $\tilde{\bm X}_{i}^k$ is the local data collecting by agent $i$ which is not allowed to exchange among agents. On the other hand,  ground truth label `0' for event $\mathcal H_0^i$ can be easily obtained by running gossip-based algorithm on $\mathcal G$.
We remark here that how to specifically set the training data collecting process is a challenging problem while it is beyond the scope of this work. Herein, we simply assume that each agent can obtain its own training data with correct labels. Other technique problems about the details of the training process will be included in another work.

\subsubsection{Temporal Difference Strategy via NN} Armed with training data, we propose a method called TDNN, which uses the time difference values as the input of the NN to perform neighborhood tasks, as illustrated in Fig. \ref{fig:nn_network}. Based on the metric in \eqref{eq:tdm}, the inputs for the two neighborhood tasks are as follows,
\beq\label{num_12}
\bm a^0=\hat{\bm a}^0= [\xi_{i1}, \xi_{i2}\cdots, \xi_{iM}]^\top.
\eeq
where $\xi_{ij}$ can be obtained by agent $i$. 
For the ND task, the computation process of NN can be described below:
\begin{align} \label{det_NN}
      {\bm a}^{h}&= \sigma ({\bm W}^{h}{\bm a}^{h-1}+{\bm b}^{h}), \quad h = 1,...,n-1;\\
       \tilde{y}_i&= g({\bm W}^{n}{\bm a}^{n-1}+{\bm b}^{n}), \quad \tilde{y}_i \overset{{\cal H}_1^{i}}{\underset{{\cal H}_0^{i}}{\gtrless}} \delta_{\mathrm {NN}},
\end{align}
where $\bm a^0$ is the input of NN. $\sigma(\cdot)$ is the activation function, $g(\cdot)$ is the sigmoid function defined as $g(\cdot)=1/(1+e^{-x})$,
$\bm W^h \in \RR^{L_h \times L_{h-1}} $ is the weight matrix between the layer $h$ and layer $h-1$, $L_h$ represents the number of neurons in layer $h$, 
$\bm b^h \in \RR^{L_h}$ and $\bm a^h \in \RR^{L_h}$ are the bias vector and the activation output in the layer $h$, respectively. 
$ \tilde{y}_i \in \RR$ is the expected output, and $\delta_{\mathrm {NN}}\in [0,1]$ is some prescribed threshold for detection task.

For the NL task, a similar NN structure is used, except for the number of neurons in the output layer. The design is given as follows,   
\begin{align} \label{loc_NN}
      \hat{\bm a}^{h}&= \sigma (\hat{\bm W}^{h}\hat{\bm a}^{h-1}+\hat{\bm b}^{h}), \quad h=1,...,n-1;\\
       \tilde{\bm z}_i&= g(\hat{\bm W}^{n}\hat{\bm a}^{n-1}+\hat{\bm b}^{n}), \quad \tilde{z}_{ij} \overset{{\cal H}_1^{ij}}{\underset{{\cal H}_0^{ij}}{\gtrless}} \epsilon_{\mathrm {NN}},
\end{align}
where $\hat{\bm a}^0$ is the input of NN, $\hat{\bm W}^{h}$, $\hat{\bm b}^{h}$, and $\hat{\bm a}^{h}$ are the weight matrix, bias term, and activation output in NN, respectively,
$\tilde{\bm z}_i=[\tilde{z}_{i1},\cdots,\tilde{z}_{iM}]\in \RR^M$ is the expected output of NL task, and $\epsilon_{\mathrm {NN}}\in [0,1]$ is some prescribed threshold.
Notice that the actual output is encoded by one-hot vector during training stage, such as $\tilde{\bm z}_i=\bm e_j$ if $j\in \mc V_m$, see Fig. \ref{fig:nn_network} (Right).

\subsubsection{Spatial Difference Strategy via NN} Both TD and TDNN only utilize the initial state and the steady state of agents rather than the transient states, leading to the possibility of losing some key features in the neighborhood tasks. In particular, the neighborhood transient state information is not effectively utilized for extracting key classification features. Therefore, we propose a strategy called SDNN to improve the detection and localization performance by using transient states and NN. As a malicious agent always tries to influence and steer the trustworthy agents away from the true value, we have
 $\EE[\bm x_j^k(t)-\overline{\bm x}_i^k(t)| \mc H_1^{ij}] \neq \EE[\bm x_j^k(t)-\overline {\bm x}_i^k(t)| \mc H_0^{ij}]$.
Thus, we can compare the state of neighbor agent $j$ and the neighborhood average of agent $i$ over time. The metrics for ND and NL tasks can be described as follows:
\beq \label{SDNN_value}
      \bm s_{ij}^k \eqdef \sum_{t=0}^{T}\big( \bm x_{j}^k(t)-  \overline{\bm x}_{i}^k(t) \big), j\in \mathcal N_i.
\eeq
\beq \label{SDNN_det}
			\chi_{ij} \eqdef \frac{1}{Kd}\sum_{k=1}^{K}  \bm 1^{\top} \bm s_{ij}^k, j\in \mathcal N_i.
\eeq
where $\bm x_{j}^k(t)$ is the $t$th states of agent $j$ at instance $k$, $\bm s_{ij}^k$ is the sum of statistical differences between agent $j$ and the neighborhood average of agent $i$. Note that $\overline{\bm x}_i^k(t)$ has been defined in \eqref{eq:avg}. 

Herein, our goal is to accurately detect insider attacks and identify whether the attacker appears in the neighborhood of agent $i$. The detection structures of SDNN are similar with that for TDNN, as seen in Fig. \ref{fig:nn_network}. Therein, we use the following inputs for these NN models of ND and NL tasks: 
\beq \label{num_3}
{\bm a}^0 = \hat{\bm a}^0 = [\chi_{i1}, \chi_{i2}, \cdots, \chi_{iM}]^\top.
\eeq
\section{Collaborative learning for a robust model} \label{sec:manner}

In previous sections, we have introduced how to use NN to help detect and localize the insider attackers. Our training data comes from a training data collecting process under Assumption \ref{ass:net} wherein the local data samples  $\tilde{\bm X}_{i}^k$ are collected by agent $i$ which could be within or beyond the neighborhood of the attacker. Apparently, the optimal train way is to upload all agents' data to a fusion center and train the model in a centralized manner. However in practice, collecting data from decentralized data sources to the center is hard due to storage and bandwidth limitations. On the other hand, as running a gossip algorithm is time-consuming, it is usually difficult and expensive to collect sufficient data at each agent. For example, the attack could occur far from the monitoring agent while the training data may only contains samples from a neighbor attacker. As the training samples may not well represent the general attack network, some individually trained NN may not fit in all insider attack events. To alleviate these issues, we propose a collaborative peer-to-peer protocol to facilitate training our NN models. Before we go to the details, we recall three assumptions for the proposed collaborative learning process. First, we assume that all agents in the network have equal number of neighbors (this is sort of impractical but we can resolve it later). Also,  different agents collect their own training data with the advantages of high scalability and privacy preservation. Moreover, we allow the trustworthy agents to have correctly labeled samples from the ND and NL tasks. For instance, the samples labeled in the current training round can be used to the next training round.
In the next, we will see how to share AI-based models between different agents to achieve robust performance in ND and NL tasks.

\subsection{The Distributed Collaborative Training Process} 
The goal of collaborative training is that participating agents acting as local learners to train good local models (i.e., NN models) through gossip exchanges.
That is, an agent $i \in \mc V$ aims to train a model that performs well with respect to the data points available on other agents.
For distributed collaborative training, the standard unconstrained empirical risk minimization problem used in machine learning problems (such as NN) can be described as follows \cite{jiang2017collaborative}:
\beq \label{problem}
 \min_{\bm W} L(\bm W)=\min \frac{1}{n} \sum_{i \in \cal V} L_i(\bm W)
\eeq
where $\bm W$ is the parameter of the NN model. $L_i(\cdot)$ is a local objective function of agent $i$, which is defined as the expected loss of the local data set. 
The local objective is to minimize the expected loss of its local sample
\beq \textstyle \label{loss}
 L_i(\bm W)=\EE_{\varsigma \sim {\cal I}_i} [\ell(\bm W, \varsigma)]
\eeq
where $\varsigma$ is a pair variable, composed of an input and related label, following the unknown probability distribution ${\cal I}_i$, which is specific for the sample set received by agent $i$.
$\ell (\cdot)$ is a loss function used to quantify the prediction error on $\varsigma$.
Let $D_i=\{\varsigma_1, \cdots,\varsigma_q \}$ represents the set of training data on agent $i \in \mc V$, which contains $q$ samples. 
Thus, we have $D =D_1\cup \cdots \cup D_n $ to optimize problem \eqref{problem}:
\begin{equation}\label{problem2}
\begin{split}
 \min_{\bm W} L(\bm W)=\min \frac{1}{n} \sum_{i\in \cal V} \Big(\frac{1}{q}\sum_{\varsigma \in D_i} \ell(\bm W, \varsigma)\Big)
\end{split}
\end{equation}
where $L_i(\bm W)=\frac{1}{q}\sum_{\varsigma \in D_i} \ell(\bm W, \varsigma)$. This formulation enables us to state the optimization problem \eqref{problem} in a distributed manner.
This distributed collaborative training problem could be addressed by gossip exchanges \cite{blot2019distributed, daily2018gossipgrad}. We will detail it as follows. 
\begin{algorithm}[t!] 
	\caption{Gossip training for AI-based methods}\label{alg:2} 
	\par\setlength\parindent{1em}\textbf{Input}:  $P_{ij}: \text{probability of exchange}$, $\eta: \text{learning rate}$.
	
\begin{algorithmic}
	\STATE{\textbf{Initialize:} $\bm W$ is initialized randomly, $i\in \mc V.$}
	\REPEAT
	\item $\bullet$  {\textsc{MergeModel}($\bm W_r, \bm W_i$) in \eqref{merge}} 
	\item $\bullet$  {$\bm W_i \leftarrow \bm W_i-\eta\hat \grd L_i(\bm W_i)$, agent $i$ updates parameters}
	\item $\bullet$  {Agent $i$ sends $\bm W_i$ to agent $j \in {\mc N}_i $ with probability $P_{ij}$}
	\UNTIL{Maximum iteration reached}
	\STATE{\textbf{function} \textsc{MergeModel}($\bm W_r, \bm W_i$)}
	
	\quad\quad {$\bm W_i \leftarrow \bm W_i(1-\mu)+ \mu \bm W_r$, $\mu \in [0, 1]$.}
	
	\STATE{\textbf{end function} }
	
\end{algorithmic} 

\end{algorithm}
\subsection{The Gossip Stochastic Gradient Descent Strategy} 
Gossip learning is a method to learn models from fully distributed data without central control \cite{hegedHus2019gossip}.
The skeleton of the gossip learning protocol is shown in Algorithm \ref{alg:2}. 
Therein, during the training stage, each agent $i$ has a NN with the same architecture and initializes a local model with the parameter $\bm W_i$. 
This is then sent to another agent $j \in \mc N_i$ in the network periodically with the probability $P_{ij}$.
Upon receiving a model $\bm W_r$, the agent $i$ merges it with the local model, and updates it using the local data set $D_i$.
We utilize the stochastic gradient descent (SGD) algorithm to estimate the local parameter $\bm W_i$ \cite{blot2018gosgd}, as follows
\beq \label{merge}
\bm W_i \leftarrow \bm W_i(1-\mu)+ \mu \bm W_r, \quad \quad \mu \in [0, 1].
\eeq
\beq \label{sgd}
  \bm W_i \leftarrow \bm W_i-\eta \hat \grd L_i(\bm W_i),  \quad \quad i\in \mc V,
\eeq
where $\eta$ and $\hat \grd L_i(\cdot)$ are the learning rate and the expected gradient of agent $i$, respectively. $\mu \in [0, 1]$ is a weight used to merge the receive model $\bm W_r$.
Herein, \textsc{MergeModel}($\bm W_r, \bm W_i$) is a merging process as shown in \eqref{merge} which is typically achieved by averaging the model with parameters, i.e., $\mu = 0.5$.

\subsection{The Tailor-degree Network} \label{sec: some_issues}
%
\begin{figure}[t]
	\begin{minipage}[b]{1.0\linewidth}
		\centering
		\centerline{\includegraphics[width=8cm]{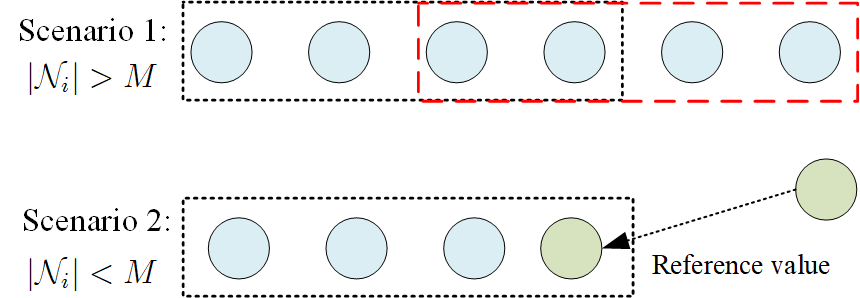}}
	\end{minipage}
	\caption{An example of tailoring neighbor agents.}
	\label{fig:degree}
\end{figure}

We have introduced the application of NN in detecting and localizing attacker, and assumed that each normal agent in the network has exactly $M$ neighbors. 
Inevitably, the communication network is an irregular network, where some agents have a heterogeneous number of neighbor agents.
In order to adapt to scenarios with different Degree-$|\mathcal N_i|$ agents, we tailor our $M$-input NN to fit into the scenario when a normal agent has $\left| \mathcal N_i \right| \neq M $ neighbors. Two scenarios are considered in this subsection. In the first scenario, we consider the case of $|{\cal N}_{i}| > M$. The $|{\cal N}_{i}|$ neighbors is divided into $\lceil |{\cal N}_i|/M \rceil$ potentially overlapping groups.
In ND and NL tasks, each group contains exactly $M$ agents, which can be treated as a standard neighbor set of TDNN and SDNN methods.
Thus, these two tasks can be implemented with the unified NN model, as seen in Fig. \ref{fig:degree}.
On the other hand, if we have $|{\cal N}_{i}| < M$, the deficient value in the input vector is replaced by a reference value to fit a Degree-$\left| {\cal N}_i \right|$ agent.
For the TDNN method, the input is reconstructed by $ {\bm a}^0 (\hat{\bm a}^0) \in {\mathbb{R}^M}= [\xi_{i1},\cdots, \xi_{i|{\cal N}_i|}, \xi_{ii}, \ldots, \xi_{ii}]^\top $,   
wherein $\xi_{ii}$ is the temporal difference value of agent $i$. 
For the SDNN method, the deficient value of input vector is replaced by $X_{ii}$, and the input is reconstructed by ${\bm a}^0 (\hat{\bm a}^0) \in {\mathbb R}^M= [\chi_{i1}; \cdots; \chi_{i{|\mathcal N_i}|};  \chi_{ii}; \cdots; \chi_{ii}]^\top $ when $|{\mc N}_{i}| < M$.

{\it Remark:}
Typically, the training data used to train the AI-based methods is collected by trustworthy agents under a scenario of specific prior information $\bm \beta$.
In practice, the prior information of the gossip-based DPS optimization protocol will be changed in some particular scenarios, that is, the test data is statistically mismatched with the training data.  
To further verify the robustness of the AI-based detection and localization models, we generate the test data by keeping the target value of attackers $\bm \alpha$ and changing the mean and the deviation of $\bm \beta$.
As depicted in Section \ref{sec:results}, we set $\bm \beta \sim \mc U [0,1]^d$, then several test scenarios are defined in the TABLE \ref{tab:1}. 
\begin{table}[t]
	\centering
	\caption{ Test scenarios settings for mismatched data}\label{tab:1}       
	\setlength{\tabcolsep}{4.7mm}
	\begin{tabular}{ccccc}
		\toprule 
		\noalign{\smallskip}
		Scenario  & Mean & Deviation & Initial distribution \\
		\noalign{\smallskip}\hline\noalign{\smallskip}
		S0 & 0.5 & 1.0 & $\bm \beta \sim$   $\mc U [0.0, 1.0]^d$\\
		S1 & 0.5 & 0.6 & $\bm \beta \sim$  $\mc U [0.2, 0.8]^d$\\
		S2 & 0.5 & 1.4 & $\bm \beta \sim$   $\mc U [-0.2, 1.2]^d$\\
		S3 & 0.7 & 1.0 & $\bm \beta \sim$   $\mc U [0.2, 1.2]^d$\\
		S4 & 0.3 & 1.0 & $\bm \beta \sim$   $\mc U [-0.2, 0.8]^d$\\
		\noalign{\smallskip}  
		\bottomrule
	\end{tabular}
\end{table}

\section{Numerical Results and Analysis} \label{sec:results} 
\begin{figure}[t!]
  \centering
	  \centerline{\includegraphics[width=6.03cm]{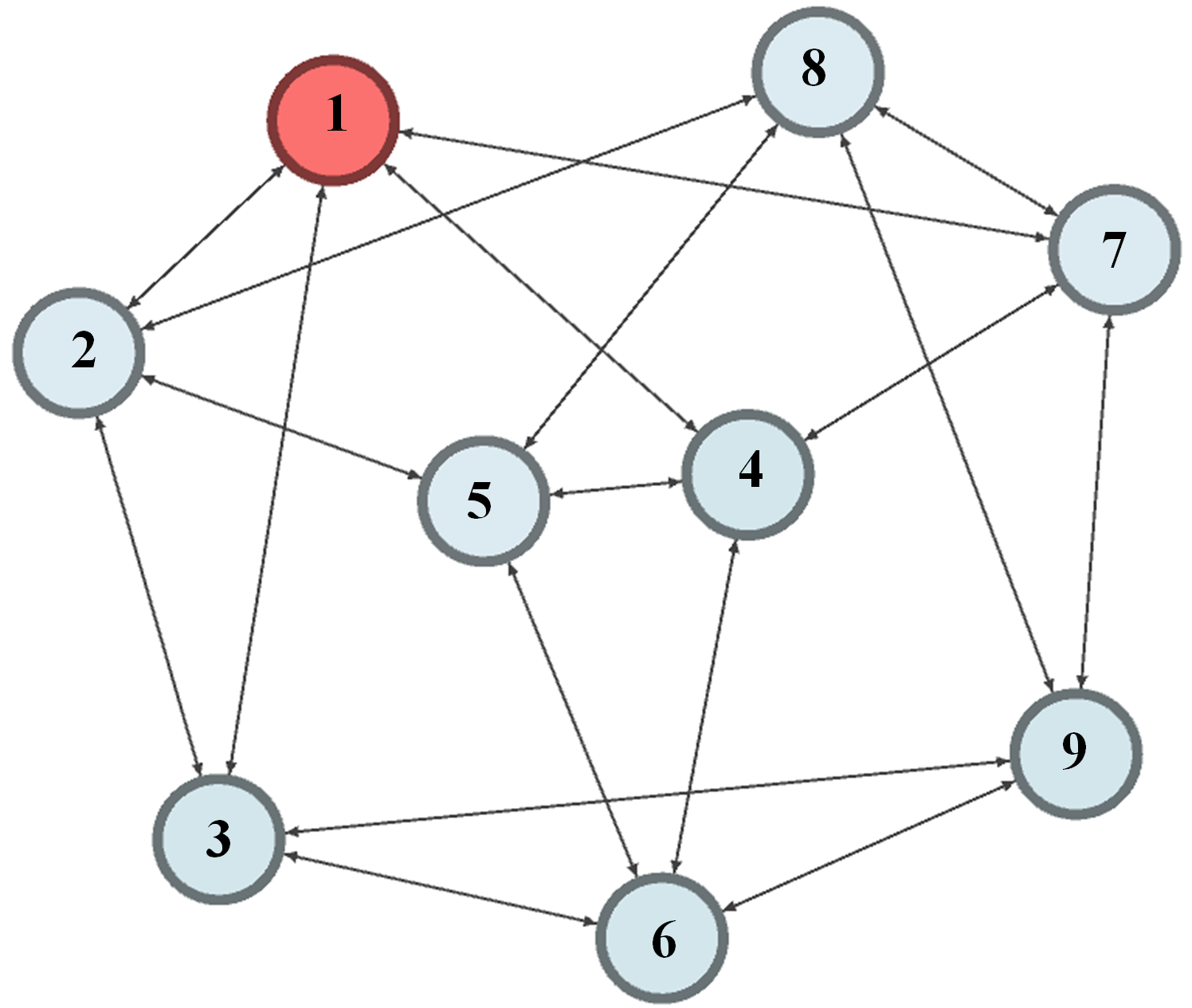}}
\caption{The Manhattan network topology with agent $1$ as the attacker.}
\label{fig:manhattan}
\end{figure}

In this section, numerical results are presented to validate the effectiveness of the proposed AI-based methods in neighborhood tasks.
The DPS algorithm runs on a Manhattan network with $n=9$ agents, as shown in Fig. \ref{fig:manhattan}.
In our experiment, an example of the least-square optimization problem is considered; i.e., in \eqref{eq:opt} we set
\begin{equation*}\label{eq.example} 
 f^k({\bm x}) = \sum_{i=1}^n f_{i}^k({\bm x})=\sum_{i=1}^n\left| ({\bm \theta}_i^k)^\top {\bm x^k} - {\phi}_i^k\right|^2, k=1,...,K.
\end{equation*} 
Herein, $f_{i}^k$ is a utility function at agent $i$.
As shown in Algorithm \ref{alg:1}, the DPS algorithm runs in an asynchronous manner such that an agent $i$ randomly selects an agent $j$ with probability $[{\bm P}]_{ij} =P_{ij} = 1 / |{\mathcal N}_i|$.
Thus the expected transition matrix in iteration $t$ can be written as ${\mathbb E}\left[ {\bm A}(t)\right]= {\bm I} - \frac{1}{2n} \bm{\Sigma} + \frac{{\bm P} + {\bm P}^\top }{2n}$, where $\bm{\Sigma}$ is a diagonal matrix with $[\bm{\Sigma}]_{ii} = \sum_{j=1}^n ({ P}_{ij} + { P}_{ji})$. 
In each instance, we set $d=2$, $T=2000$, the initialization ${\bm x}^k(0) \sim {\mathcal U}[0,1]^d$, ${\bm \alpha}^k \sim {\mathcal U}[-0.5,0.5]^d$ and $\bm r_j^k(t) \sim {\mathcal U}[-\hat{\lambda}^t,\hat{\lambda}^t]$, where $\lambda$ is the second largest eigenvalue of $\EE[{\bm A}(t)]$.
In particular, to serve our purpose we change the function $f_i^k({\bm x})$ by randomly generating ${\bm \theta}_i^k \sim {\mathcal U}[0.5,2.5]^d$, $({\bm x}^\star)^k \sim {\mathcal U}[0,1]^d$, and thus we have $\phi_i^k=(\bm \theta_i^k)^\top(\bm x^\star)^k$.

For the AI-based methods, the feed forward neural networks (FFNN) with three hidden layers is applied to perform the ND and NL tasks with neurons in each hidden layer being $200$, $100$ and $50$ respectively. These NNs are implemented using a modified version of the deep learning toolbox in \cite{palm2012prediction}. Rectified linear unit (ReLU) as the activation function is equipped in all hidden layers and the parameters of NN are jointly optimized through a back propagation method by minimizing the loss function defined on different tasks.

To provide neighbor data and ground truth labels for the AI-based methods, we run the DPS algorithm independently in each event \eqref{ND_task}, starting with a new initial state each time. As the Manhattan network is symmetric, to obtain the training data under hypothesis $\mathcal H_1^i$ with label `1', we need to collect data at two types of the trustworthy agents: the one that stands at the position ``next to" the attacker agent (for example agent $1$ is the only attacker and we collect data at agent $2$), and the one that stands at the position ``far from" the attacker agent (for example agent $1$ is the only attacker and we collect data at agent $5$). Meanwhile, the training data under hypothesis $\mathcal H_0^i$ with label `0' is collected at any agent when DPS is running on the Manhattan network free of attacker. We collect data from different scenarios as shown in Table~\ref{tab:2} and fuse
them into ND and NL models. Therein, available data is typically split into two sets, a training set and a testing set.
As for the ND task, the detection task consists of $30,000$ samples as the training set and $18,000$ samples as the testing set.  Herein, within the data under hypothesis $\mathcal H_1^i$, we have $10000$ samples collecting at agent next to the attacker and $10000$ samples collecting at agent far from the attacker.
For the NL task, the training set and testing set contain $10,000$ and $6,000$ samples respectively. Herein, we encode the ground truth labels of event $\mathcal H_1^i= \mathcal H_1^{ij} \cup \mathcal H_0^{ij}$ by one-hot coding, where the neighbor attacker is labeled by `1' and the trustworthy agent by `0'.
\begin{table}[t]
	\centering
	\caption{ Training and testing sets for AI-based methods given thatz agent $j$ is the attacker. }\label{tab:2}       
	\begin{tabular}{c c c c c c c}
		\toprule 
		\noalign{\smallskip}
		Task  & Event & Training set & Testing set & Label \\ 
		\noalign{\smallskip}\hline\noalign{\smallskip}
		\multirow{5}{*}{ND} & $\mc H_0^i$ & 10000 & 6000 & 0 \\ 
		\noalign{\smallskip}\cline{2-5}\noalign{\smallskip}
		& $\mc H_{1}^i ( j \in \mc N_i)$ & 10000 & 6000 & 1 \\ 
		\noalign{\smallskip}\cline{2-5}\noalign{\smallskip}
		& $\mc H_{1}^i ( j \notin \mc N_i)$ & 10000  & 6000 & 1 \\
		\noalign{\smallskip}\hline\noalign{\smallskip}
		NL & $\mc H_1^i$ & 10000 & 6000 & $\bm e_j$ \\ 
		\noalign{\smallskip} 
		\bottomrule
	\end{tabular}
\end{table}

Usually, the detection and localization models of AI-based methods are actually classifiers for which NN produces continuous quantities to predict class membership through different thresholds.
To make a more comprehensive evaluation of these classifiers, we adopt the probabilities of detection and false alarm for the ND and NL tasks. That is, we define
\begin{align}
     P_{nd}^i \eqdef P(\hat{\mathcal H}^i=&\mathcal H_1^i | \mathcal H_1^i),~P_{nf}^i \eqdef P(\hat{\mathcal H}^i=\mathcal H_1^i | \mathcal H_0^i). \\
		 P_{ld}^i \eqdef P(\hat{\mathcal H}^{ij}=&\mathcal H_1^{ij} | \mathcal H_1^{ij}), ~P_{lf}^{i} \eqdef P(\hat{\mathcal H}^{ij}=\mathcal H_1^{ij} | \mathcal H_0^{ij}). 
\end{align}
where $\hat{\mathcal H}^i$ and $\hat{\mathcal H}^{ij}$ are the estimated event based on AI-based methods.
$P_{nd}^i$ ($P_{ld}^i$) and $P_{nf}^i$ ($P_{lf}^{i}$) are the probability of detection and false alarm in the ND (NL) task, respectively. 
More specifically, those probabilities are calculated as follows
\begin{align} 
         P_{nd}^i=&\frac{1}{N_{nd}}\sum_{n=1}^{N_{nd}}I(y_{i}^{(n)}=\hat{y}_{i}^{(n)}=1), \\
         P_{nf}^i=&\frac{1}{N_{nf}}\sum_{n=1}^{N_{nf}}I(y_{i}^{(n)}=0 \wedge \hat{y}_{i}^{(n)}=1),\\
			   P_{ld}^i=&\frac{1}{N_{ld}}\sum_{n=1}^{N_{ld}}I(z^{(n)}_{ij}=\hat{z}^{(n)}_{ij}=1),\\
       	 P_{lf}^i=&\frac{1}{N_{lf}}\sum_{n=1}^{N_{lf}}I(z^{(n)}_{ij}=0 \wedge \hat{z}^{(n)}_{ij}=1).
\end{align}
where $N_{nd}$ ($N_{nf}$) and $N_{ld}$ ($N_{lf}$) are the number of positive (negative) samples in the ND and NL tasks, respectively, 
$\hat{y}_{i}^{(n)}$ ($\hat{z}^{(n)}_{ij}$) is the predicted class by ND (NL) classifiers, and $y_{i}^{(n)}$ ($z^{(n)}_{ij}$) is the ground-truth class label.
Note that $I(\cdot)$ is an indicator function that has the value $1$ when the predicted class label equal to the ground-truth class label.
Based on these probabilities, the detection (or localization) performance can be investigated by the receiver operating characteristic (ROC)\cite{fawcett2006introduction}, for which the probability of detection is plotted on the $Y$-axis and the probability of false alarm is plotted on the $X$-axis.
It is worth noting that the ROC curves that approach the upper left corner outperform those far from it.
\subsection{Detection and Localization for One attacker} \label{sec: one_attacker}
\begin{figure}[t!]
\begin{minipage}[b]{.49\linewidth}
  \centering
  \centerline{\includegraphics[width=4cm]{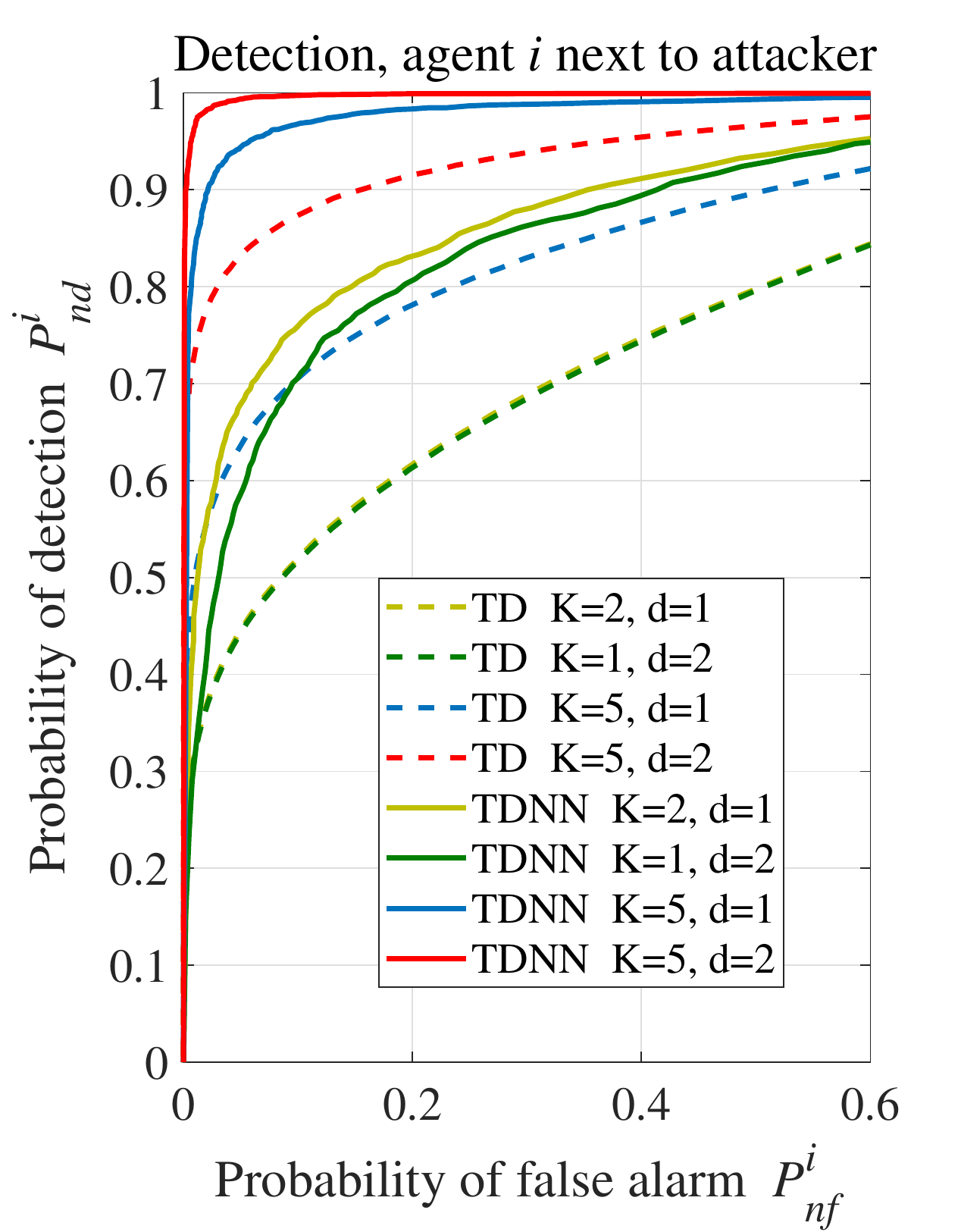}} 
\end{minipage}
\hfill
\begin{minipage}[b]{0.49\linewidth}
  \centering
  \centerline{\includegraphics[width=4cm]{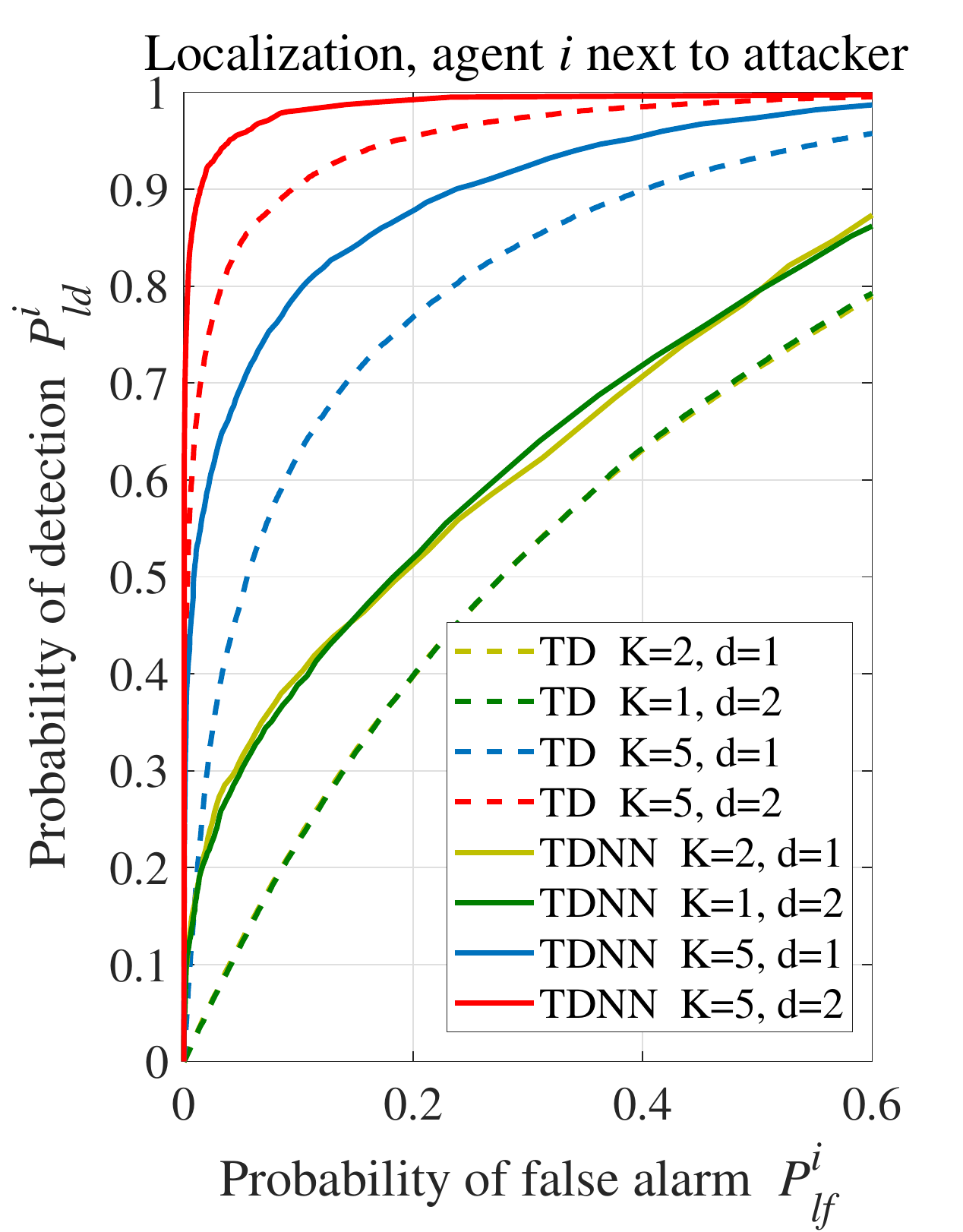}}
\end{minipage}
\caption{ROCs of TDNN and TD methods: (Left) ND task, (Right) NL task.}
\label{fig:ROC_one_TDNN}
\end{figure}

\begin{figure}[t!]
\begin{minipage}[b]{.49\linewidth}
  \centering
  \centerline{\includegraphics[width=4cm]{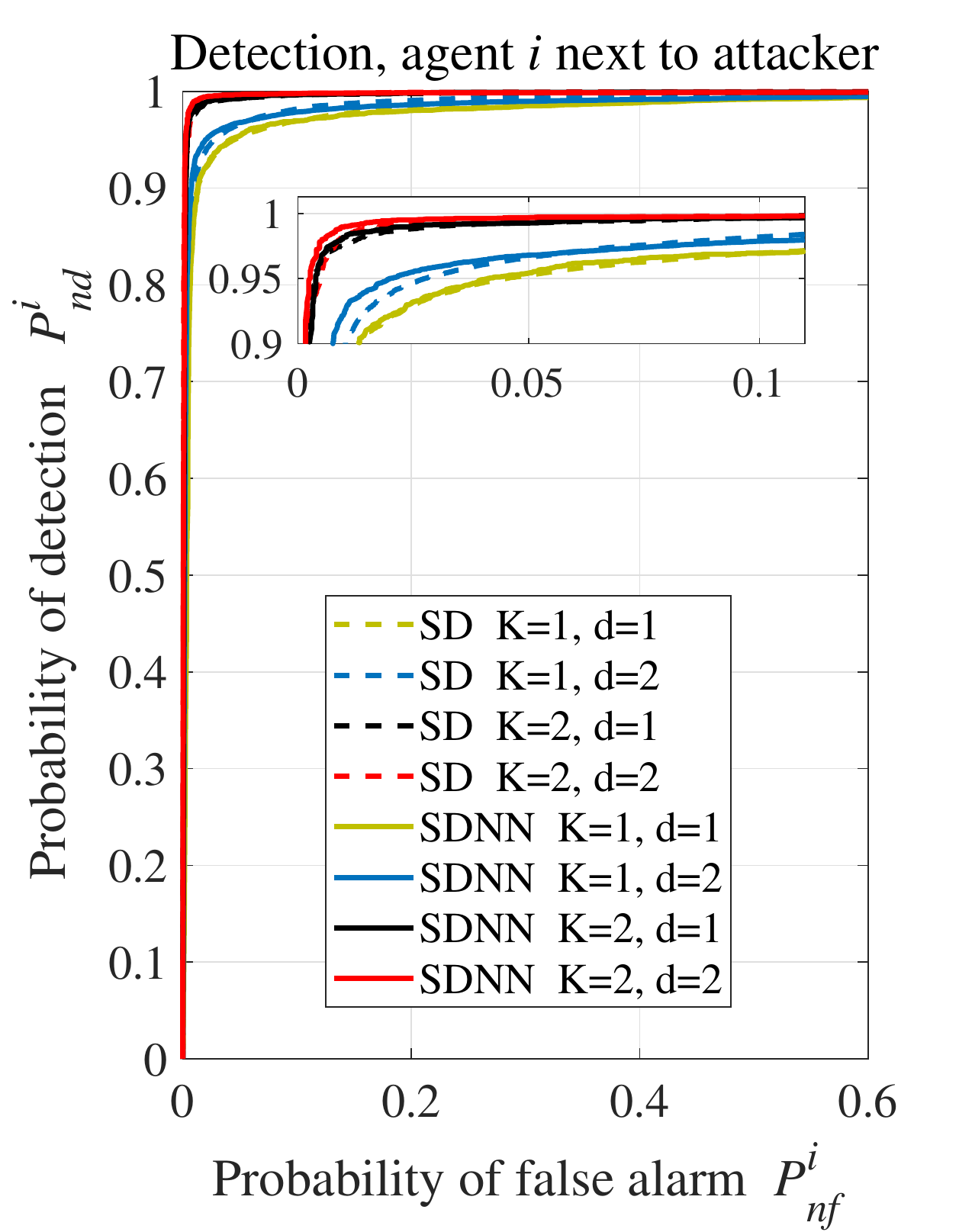}} 
\end{minipage}
\hfill
\begin{minipage}[b]{0.49\linewidth}
  \centering
  \centerline{\includegraphics[width=4cm]{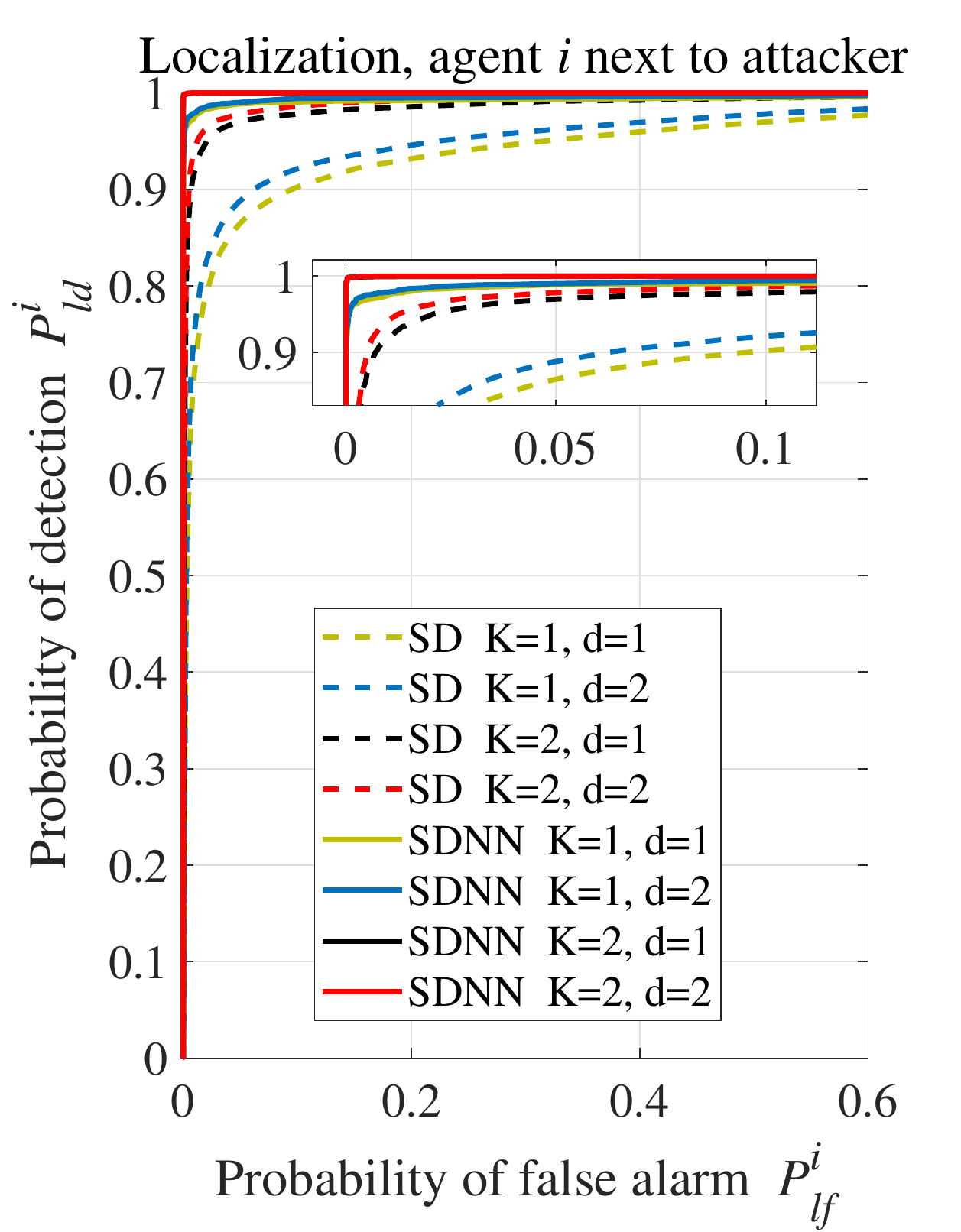}}
\end{minipage}
\caption{ROCs of SDNN and SD methods: (Left) ND task, (Right) NL task.}
\label{fig:ROC_one_SDNN}
\end{figure}
In this subsection, we show the detection and localization performance of AI-based methods when the Manhattan network contains only one attacker, seen in Fig. \ref{fig:manhattan}. Suppose that agent $1$ is the attacker and the monitoring node is agent $2$ (or $3$, $4$ and $7$). Then, the NN model is trained and also tested by the data collecting from the agent next to the attacker.

In Fig. \ref{fig:ROC_one_TDNN}, we study the attacker detection and localization performance of TDNN, where TD in \cite {icasspsttackerWu} is taken as the benchmark method. 
ROC curves of ND task is depicted in Fig. \ref{fig:ROC_one_TDNN} (Left), variables $K$ and $d$ in the legend are the number of instances and dimensions used to detect insider attacks respectively.
The localization performance of NL task is shown in Fig. \ref{fig:ROC_one_TDNN} (Right), where we assume that the ND task can completely distinguish between events $\mc H_0^i$ and $\mc H_1^i$ without errors (by an `Oracle'). In these plots, it is obvious that both the performance of TDNN and TD will improve significantly as $K$ increases when $d$ is fixed, and vice versa. For the first and second curves in ND and NL tasks, TD 
has the same performance at $K=2, d=1$ as at $K=1, d=2$, and the same is true for TDNN on the fifth and sixth curves, which is inherent to TD strategy \eqref{eq:tdm}.
Thus, we may say that either increasing $K$ or increasing  $d$ will bring the same improvement over performance.
From Fig. \ref{fig:ROC_one_TDNN}, it can be seen that TDNN improves significantly over TD in terms of both detection and localization performance, performing good performance when $K=5, d=2$. 

The ROCs of SDNN are shown in Fig. \ref{fig:ROC_one_SDNN}, while SD in \cite{li2020detect} is selected as a benchmark.
It can be seen from the plots that both SDNN and SD already provide good detection and localization performance when $K=2$, which is better than that of TDNN and TD methods.
This result implies that transient states can indeed provide more information to identify the attacker, as spatial methods (SDNN and SD) leverage the entire dynamic information while the temporal methods (TDNN and TD) only utilize the first and last states. 
Also in this case, the attacker detection and localization performance of SDNN and SD will improve significantly as $K$ increases when $d$ is fixed. 
When $K$ is fixed, the performance of SDNN and SD slightly improved as $d$ increases.
For the ND task in Fig. \ref{fig:ROC_one_SDNN} (Left), the detection performance of SDNN and SD are close to each other, they show excellent performance under the same feature processing condition, seen in \eqref{eq:sdm} and \eqref{SDNN_value}. 
Nevertheless, SDNN has a drastic advantage over SD method in NL task and can completely distinguish the neighboring attacker at $K=2$, seen in Fig. \ref{fig:ROC_one_SDNN} (Right).

\subsection{Performance of the Collaborative Learning}


\begin{figure}[t!]
	\begin{minipage}[b]{.49\linewidth}
		\centering
		\centerline{\includegraphics[width=4cm]{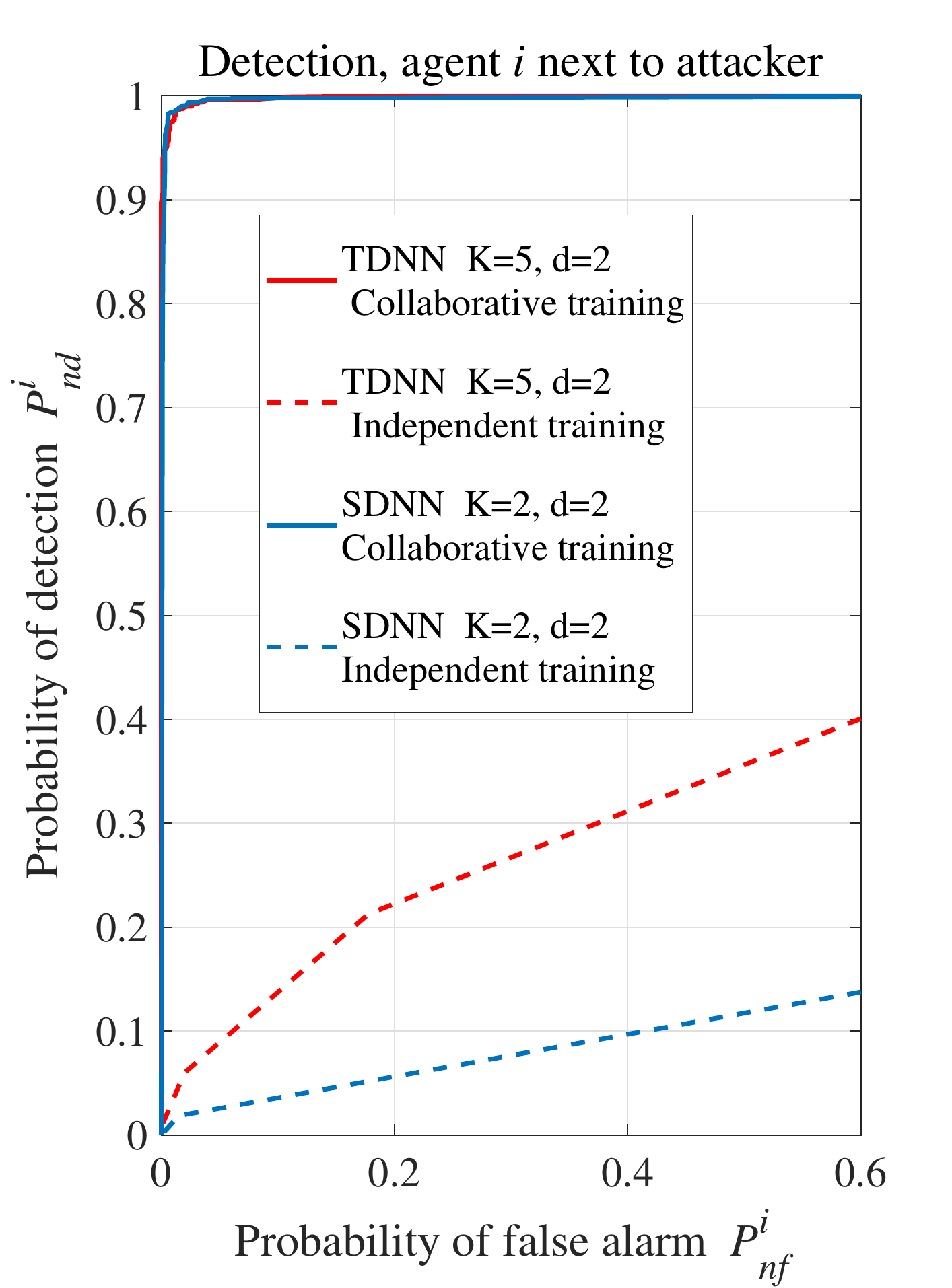}} 
	\end{minipage}
	\hfill
	\begin{minipage}[b]{0.49\linewidth}
		\centering
		\centerline{\includegraphics[width=4cm]{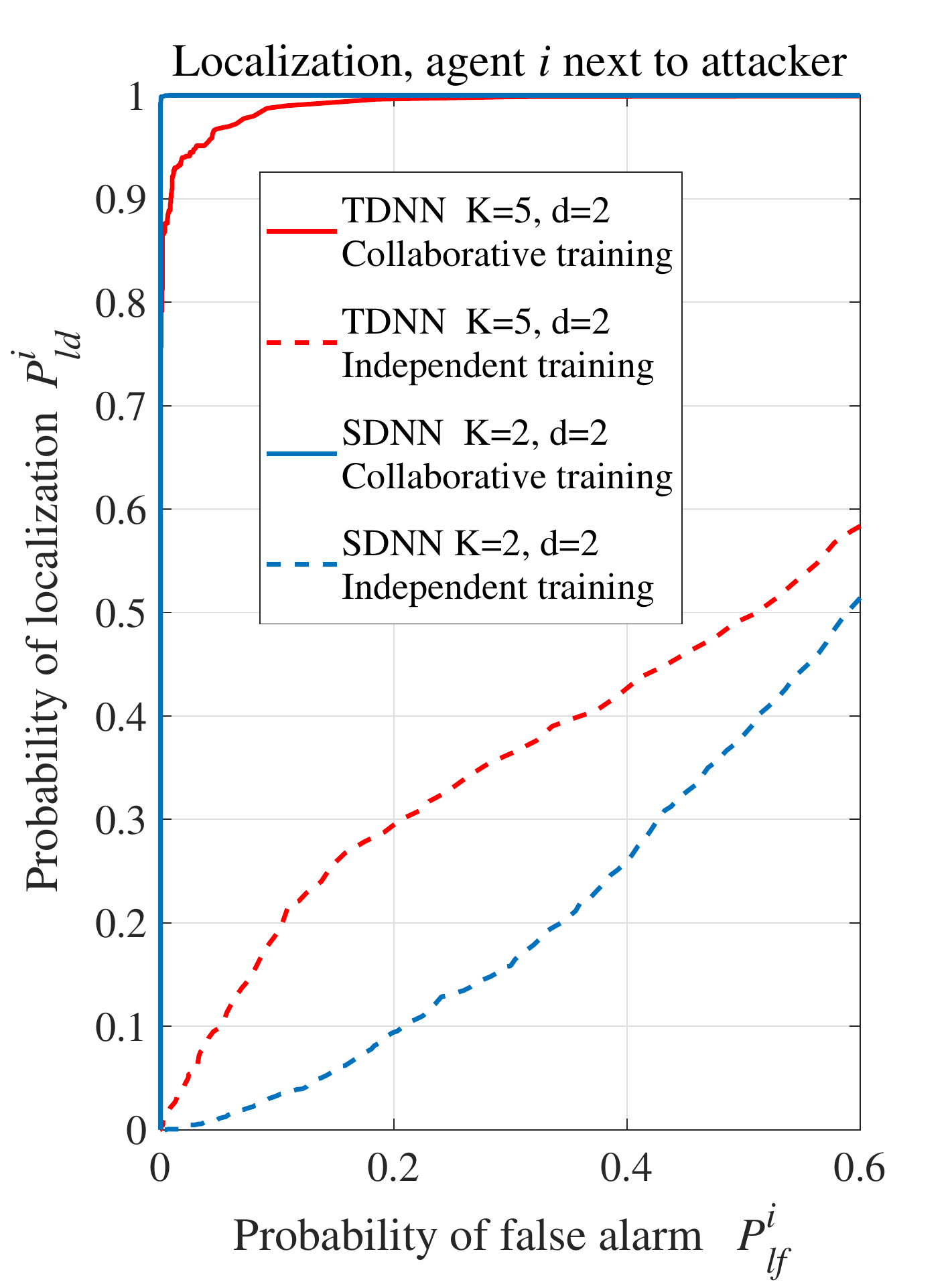}}
	\end{minipage}
	\caption{Comparison between independent training and collaborative training based on matched data. Models are trained on sufficient ``next to" data then tested on ``next to" data.}
	\label{fig:gossip_merge}
\end{figure}

\begin{figure}[t!]
	\begin{minipage}[b]{.49\linewidth}
		\centering
		\centerline{\includegraphics[width=4cm]{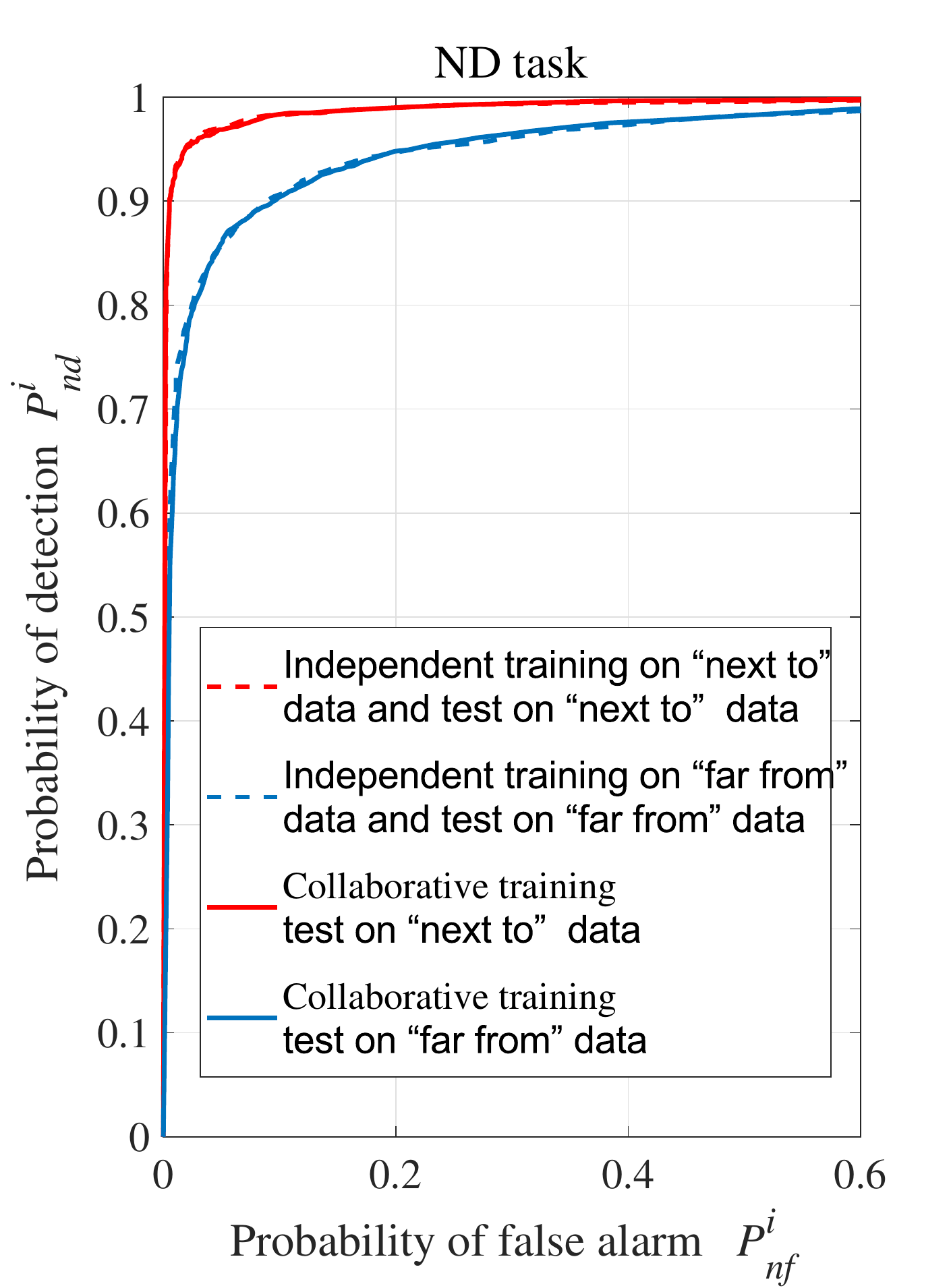}} 
	\end{minipage}
	\hfill
	\begin{minipage}[b]{0.49\linewidth}
		\centering
		\centerline{\includegraphics[width=4cm]{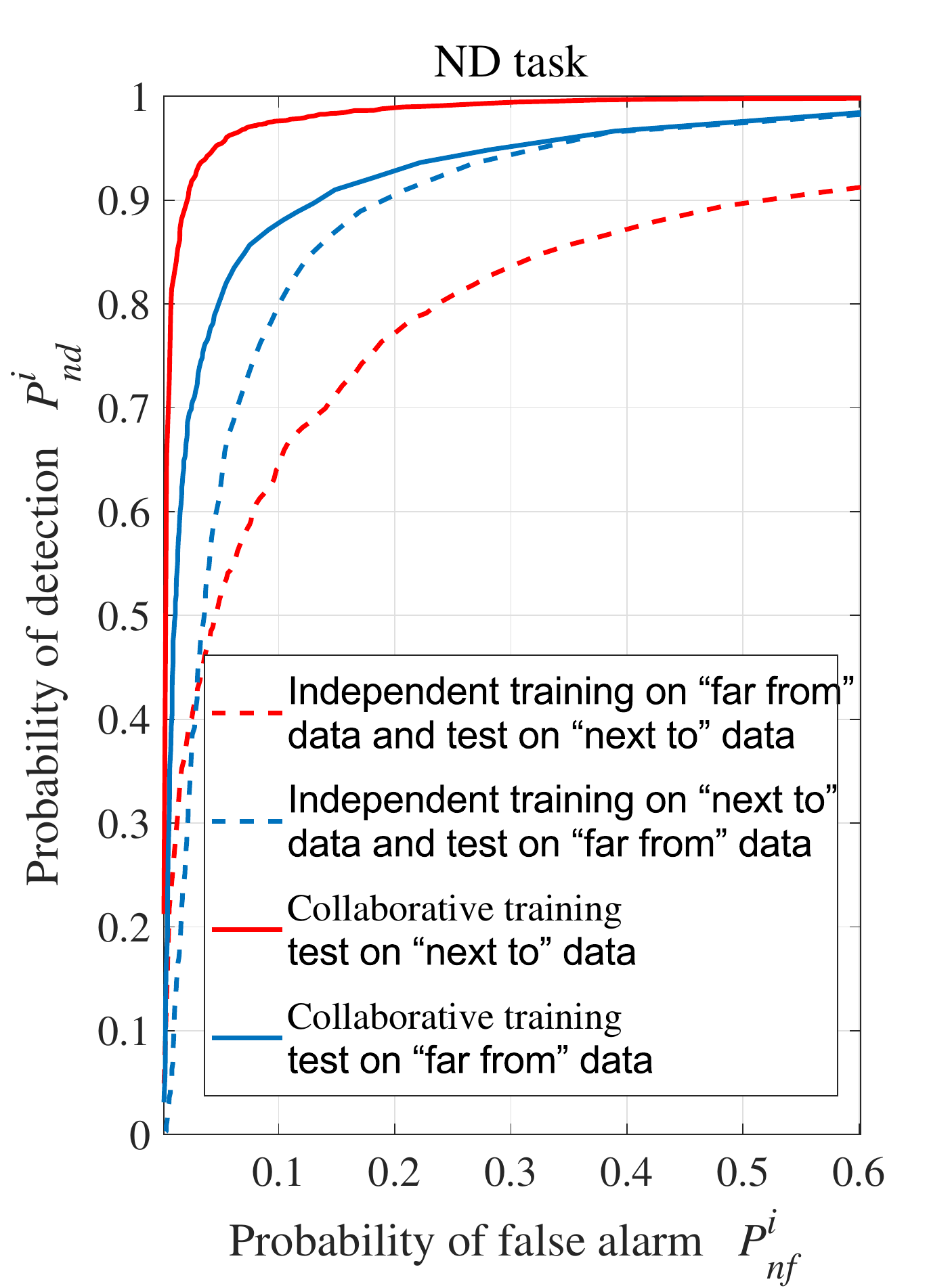}}
	\end{minipage}
	\caption{Comparison between independent training and collaborative training for mismatched data. The plots are SDNN with $K=1, d=2$.
	}
	\label{fig:farFrom_nextTo}
\end{figure}

In this subsection, we show how to utilize the collaborative learning protocol to train a robust model for accommodating more attack events. Specifically, we consider the performance comparison of independent training and collaborative training. Herein, independent training means that each agent trains its model based on its local data, where ``next to" data refers to samples collected at the agent next to an attacker while ``far from" data refers to samples collected at the agent far from an attacker. Moreover, ``next to" model refers as the independent model trained on ``next to" data while ``far from" model refers as the independent model trained on ``far from" data. We then consider two extreme cases to verify the collaborative learning.

For Case $1$, we assume that in the independent training process the monitoring agent only collects a very small amount of local ``next to" data, which is not enough to complete a meaningful NN model.
While in the collaborative training process, the agent update its local model by merging the received neighboring models.  We then test the two training methods on ``next to" data for ND and NL tasks. In Fig. \ref{fig:gossip_merge}, the dashed and solid lines are the performance for the independent training and the collaborative training, respectively.
It is clear that with insufficient samples, the agent in independent learning performs poorly on both ND and NL tasks, while
the collaborative training enables the agent to learn models from its neighbors, which greatly improves detection and localization performances. It is expected that a similar result also holds for ``far from" cases. 

For Case $2$, we first consider the scenario that each agent has sufficient ``next to" data (``far from" data) to train the NN model. We then test the independent/collaborative training models on the ``next to" data (``far from" data), which is matched the training data. The red (blue) dashed and solid lines in Fig. \ref{fig:farFrom_nextTo} (Left) represent their ROC results, which imply that the collaborative learning model will converge to the independent learning model. On the other hand, we further test the mismatch case of the training data and testing data. That is, we use the ``next to" data to test the ``far from" model, and vice versa. Interestingly, Fig. \ref{fig:farFrom_nextTo} (Right) shows that the collaborative learning model has a significant improvement over the independent model, as it learns both the characteristics of the ``next to" model and ``far from" model.
These results demonstrate the advantage of the collaborative learning on the robustness of the model. Therefore, when there are enough samples, collaborative learning also has strong competitiveness compared with independent learning.

\subsection{Performance for Different Degree-\text{$|\mc N_i|$} agents}
\begin{figure}[t!]
	\begin{minipage}[b]{.49\linewidth}
		\centering
		\centerline{\includegraphics[width=4cm]{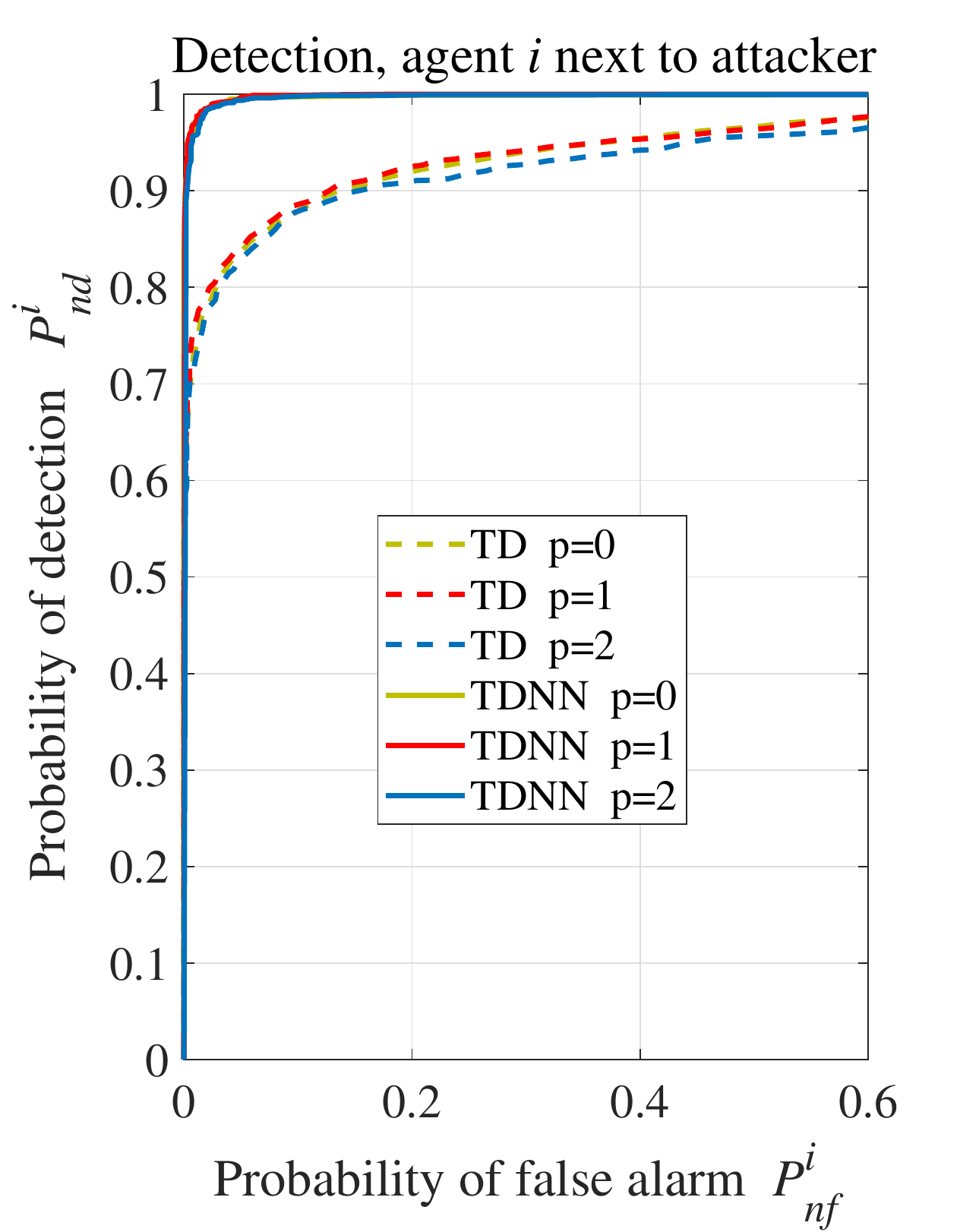}} 
	\end{minipage}
	\hfill
	\begin{minipage}[b]{0.49\linewidth}
		\centering
		\centerline{\includegraphics[width=4cm]{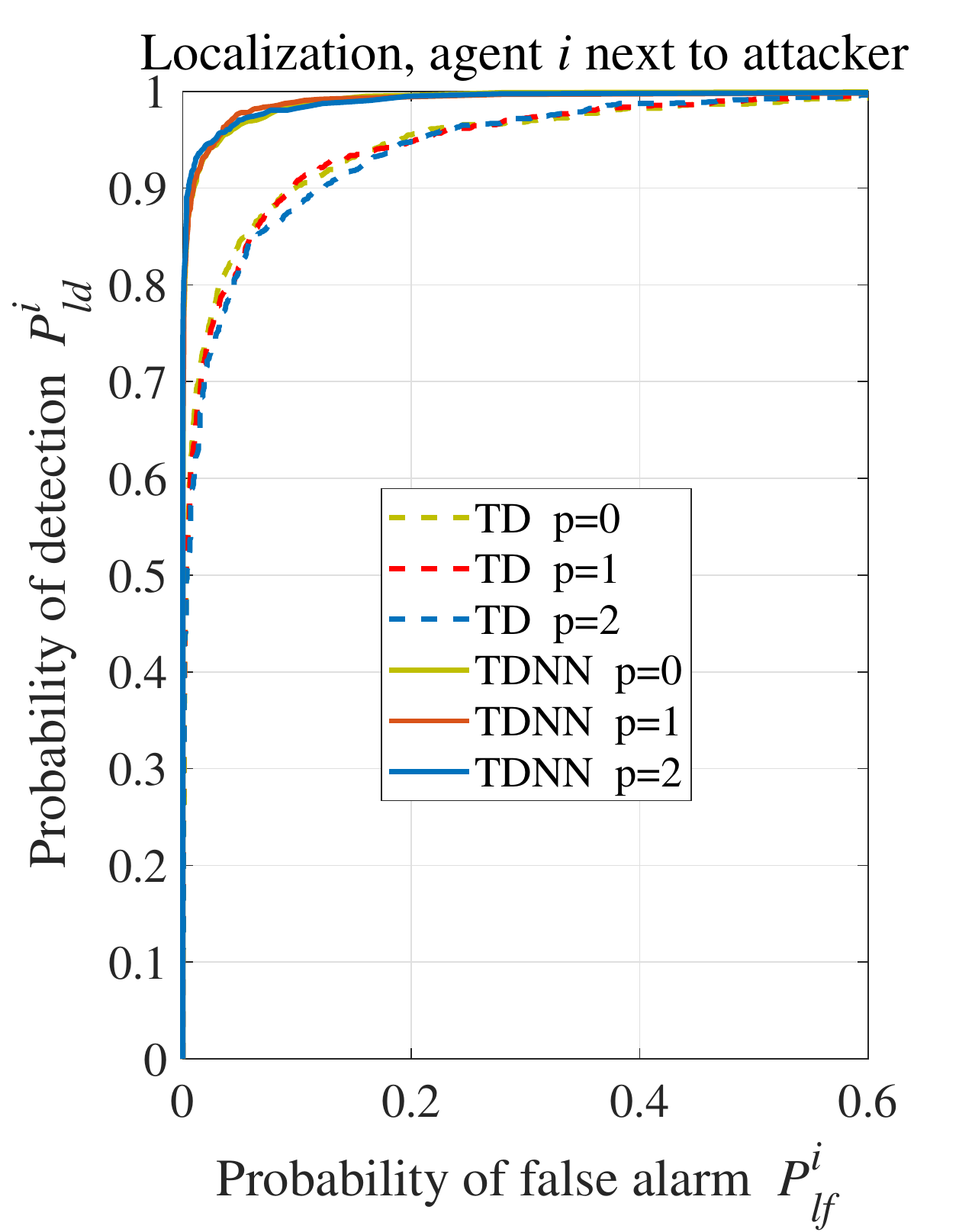}}
	\end{minipage}
	\caption{ROCs of TDNN and TD with different deficient size: (Left) ND task, (Right) NL task. $p={M- |{\cal N}_i|}$ means that the number of deficient inputs.}
	\label{fig:ROC_degree_TDNN}
\end{figure}

\begin{figure}[t!]
	\begin{minipage}[b]{.49\linewidth}
		\centering
		\centerline{\includegraphics[width=4cm]{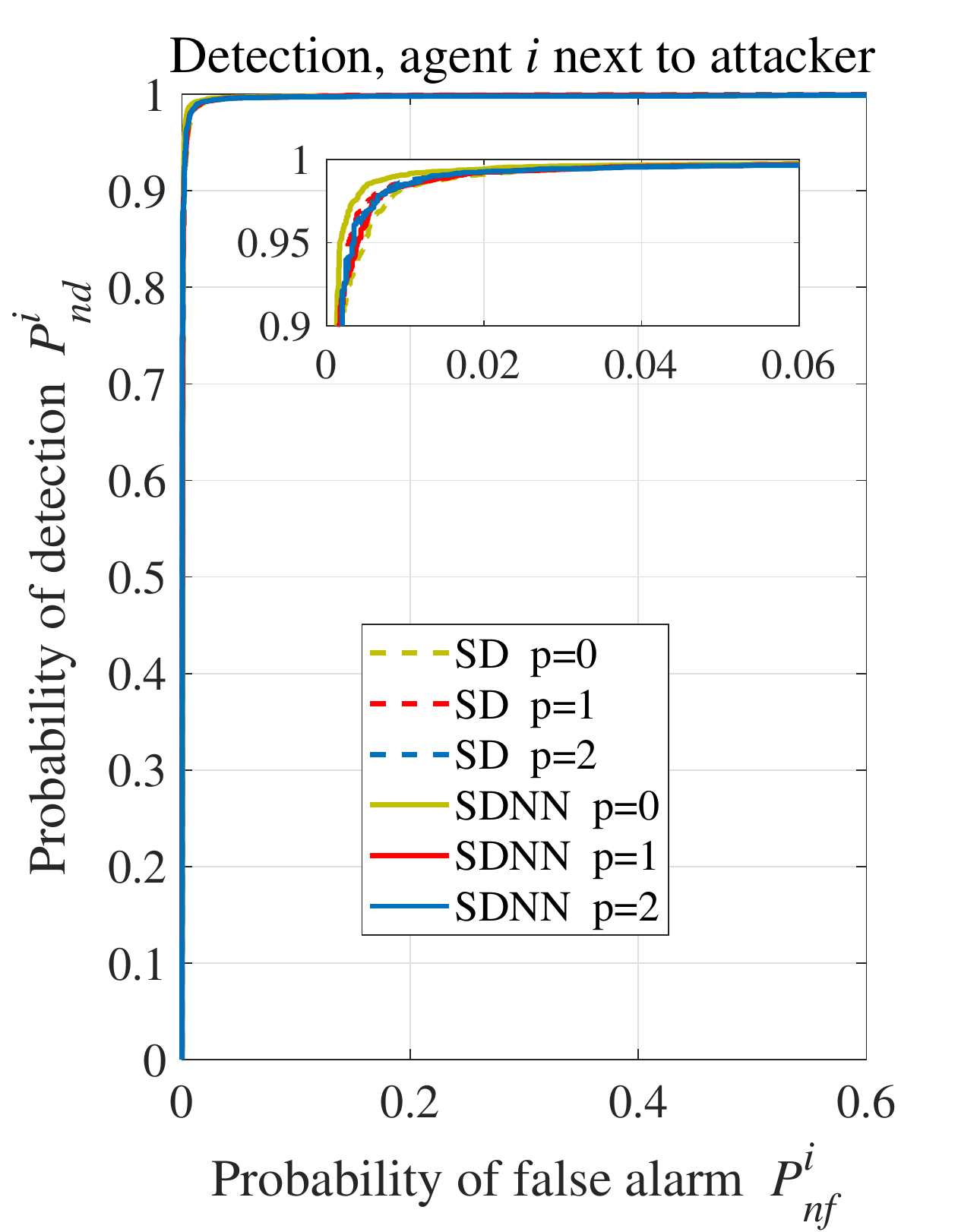}} 
	\end{minipage}
	\hfill
	\begin{minipage}[b]{0.49\linewidth}
		\centering
		\centerline{\includegraphics[width=4cm]{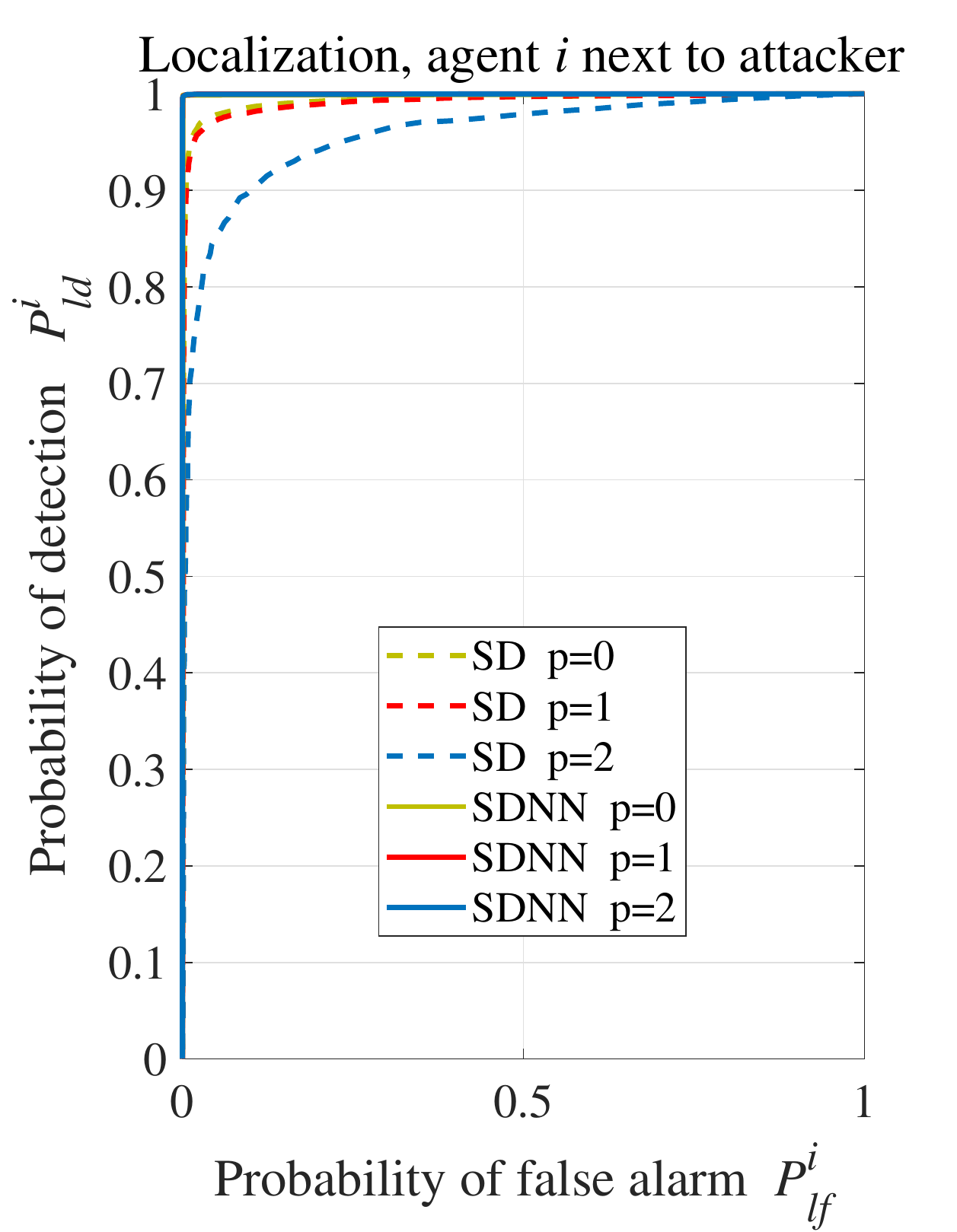}}
	\end{minipage}
	\caption{ROCs of SDNN and SD with different deficient size: (Left) ND task, (Right) NL task. $p={M- |{\cal N}_i|}$ means that the number of deficient inputs.}
	\label{fig:ROC_degree_SDNN}
\end{figure}
In this subsection, we discuss the scenario where the communication network is an irregular network. We assume that the number of mismatched inputs not matching the unified model is $p={M- |{\cal N}_i|}$.
According to the scheme described in subsection \ref{sec: some_issues}, we test the detection and localization performance of AI-based methods when $p \neq 0$. 
To set up this simulation, the Manhattan network topology is selected as the target example, and we have $M=4$ and $|{\cal N}_i|=\{2, 3\}$.
The attack scenario of $m=1, c=1$ is applied to verify our proposed method, and the performance in $p=0$ is taken as the baseline.
We choose agent $2$ as the test agent whose neighbors are agents $1, 3, 5$ and $8$, i.e., $p=0$.
When $p=1$, we cut off the connection between agents $2$ and $3$, then the neighbors of the test agent are agents $1$, $5$, and $8$.
Next when $p=2$, we further cut off the connection between agents $2$ and $5$, leaving only agents $1$ and $8$ as neighbors of the test agent $2$.
Note that the parameters for the AI-based methods are the same as those in previous subsection \ref{sec: one_attacker}, and that the testing data is generated from a modified Manhattan network with $p=1$ and $p=2$.

Fig. \ref{fig:ROC_degree_TDNN} shows the attacker detection and localization performance of TDNN and TD methods with $K=5, d=2$. 
It can be seen that the performance of TDNN and TD in ND and NL tasks will not fluctuate significantly with the increase of $p$.
However, TDNN has more stable detection and localization performance than TD.
In Fig. \ref{fig:ROC_degree_SDNN}, we shows the performance of SDNN and SD methods with $K=2, d=2$.
As $p$ increases, SD has good detection performance, but its localization performance slightly decreases.
Obviously, the results in both ND and NL tasks show that in our setting $p$ does not have a significant effect on the performance of SDNN.
SDNN still can provide stable performance for detecting and localizing attackers.
These results suggest that the proposed AI-based models may fit well with irregular degree networks.

\subsection{Robustness test when data is mismatched with prior information}
\begin{figure}[t!]
	\begin{minipage}[b]{.49\linewidth}
		\centering
		\centerline{\includegraphics[width=4cm]{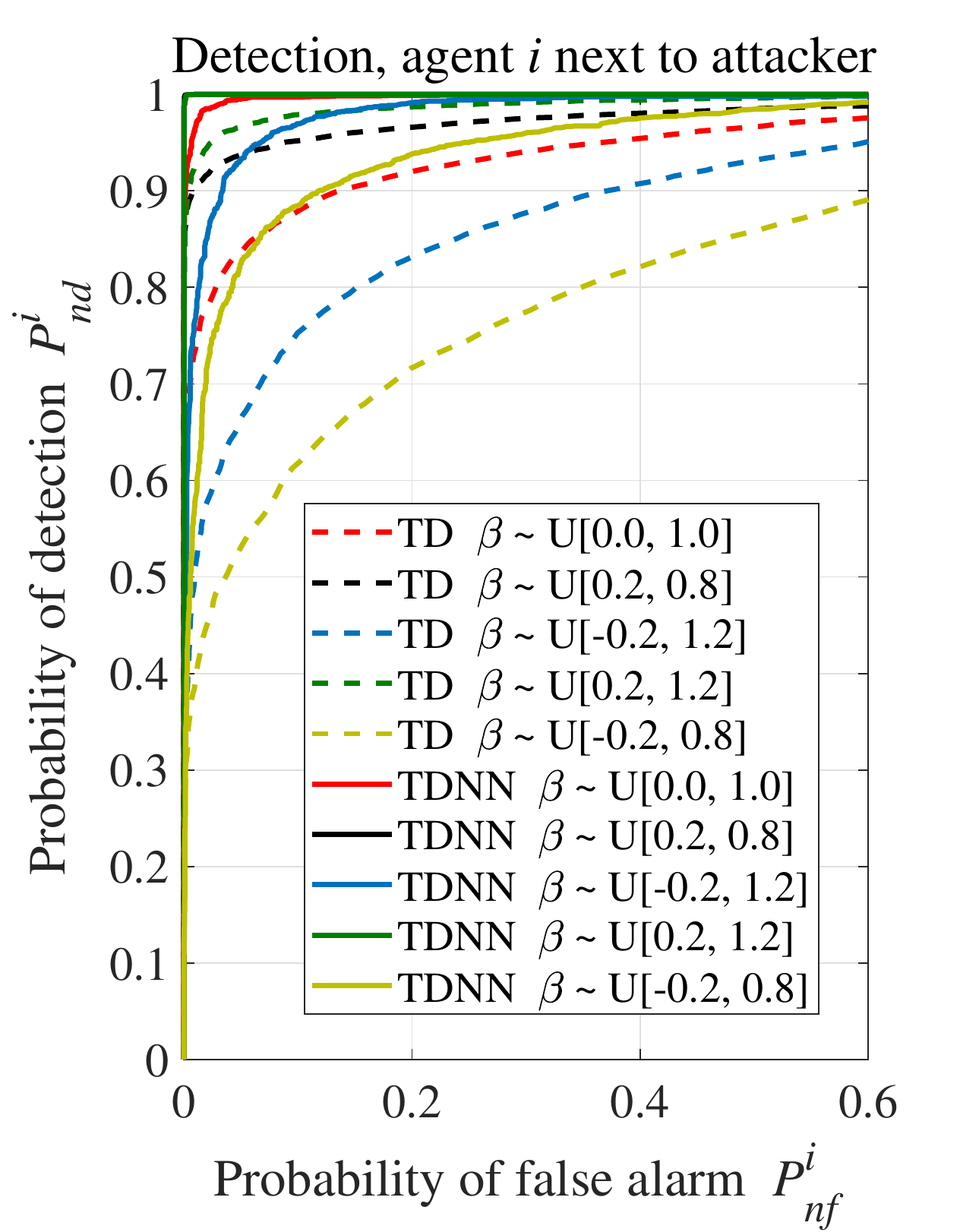}} 
	\end{minipage}
	\hfill
	\begin{minipage}[b]{0.49\linewidth}
		\centering
		\centerline{\includegraphics[width=4cm]{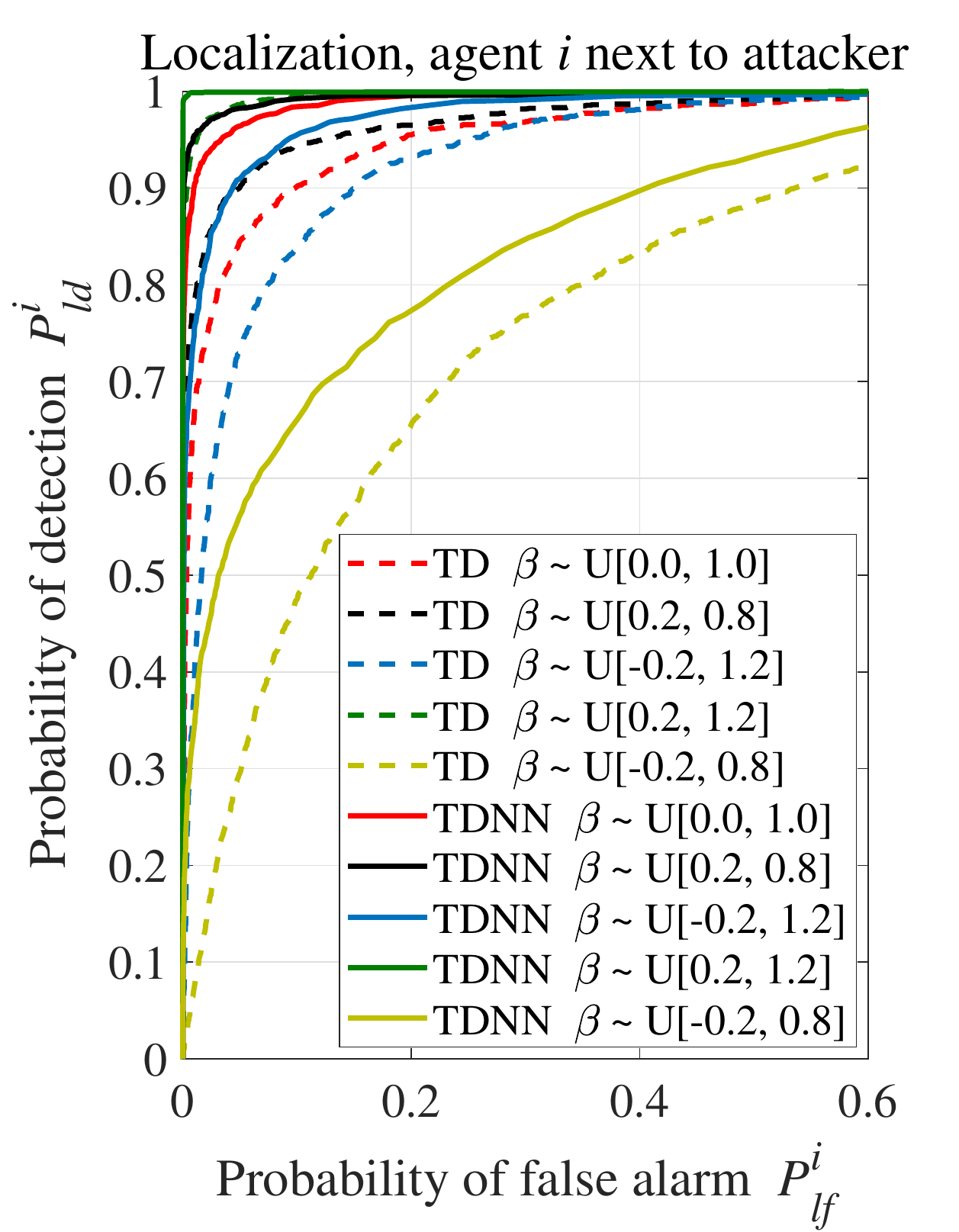}}
	\end{minipage}
	\caption{ROCs of TDNN and TD for the mismatch model: (Left) ND task, (Right) NL task. ${\bm \alpha}^k \sim {\mathcal U}[-0.5,0.5]^d$. Each entry of ${\bm x}^k(0)$ is distributed as legended for testing data.}
	\label{fig:ROC_robust_TDNN}
\end{figure}
\begin{figure}[t!]
	\begin{minipage}[b]{.49\linewidth}
		\centering
		\centerline{\includegraphics[width=4cm]{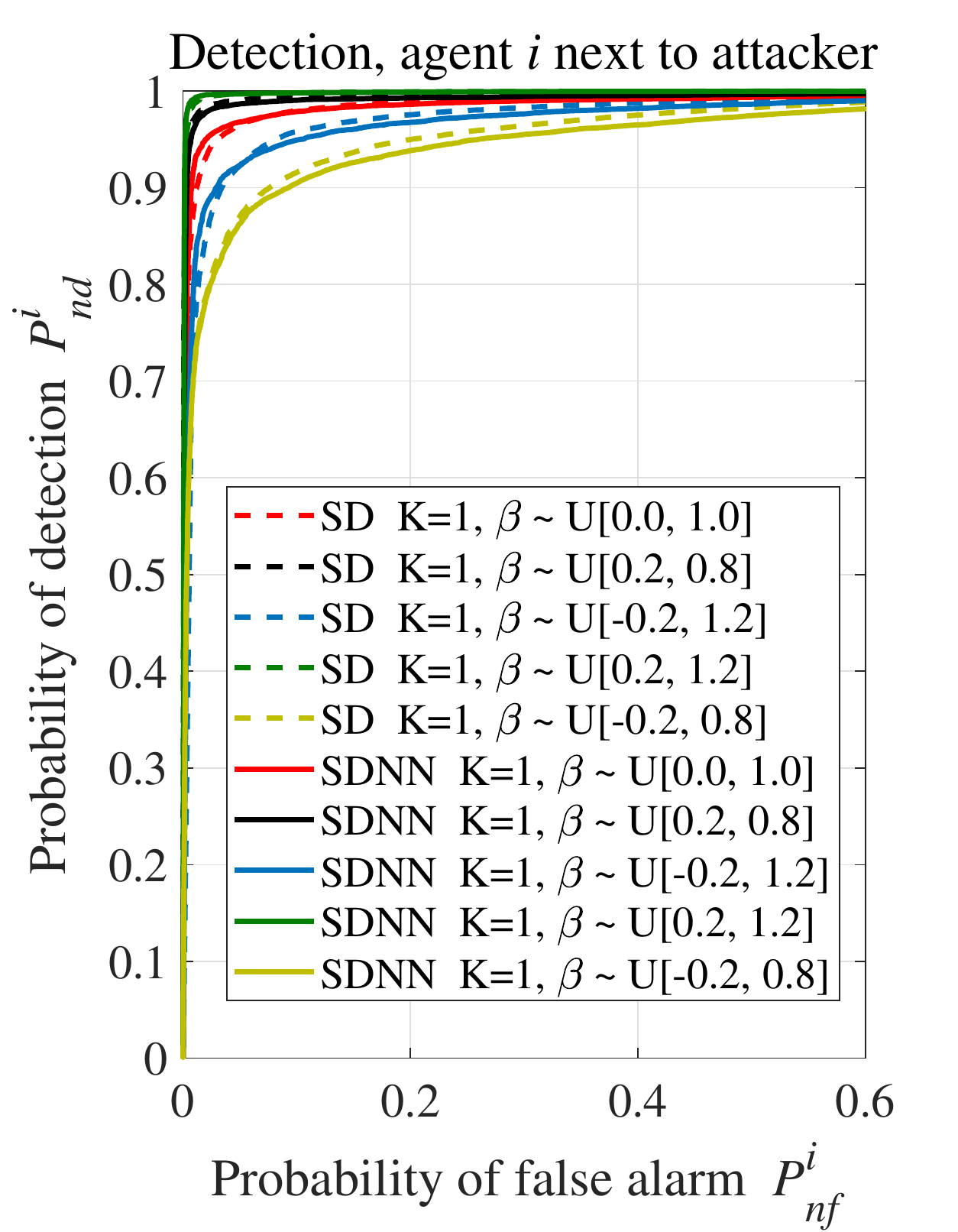}} 
	\end{minipage}
	\hfill
	\begin{minipage}[b]{0.49\linewidth}
		\centering
		\centerline{\includegraphics[width=4cm]{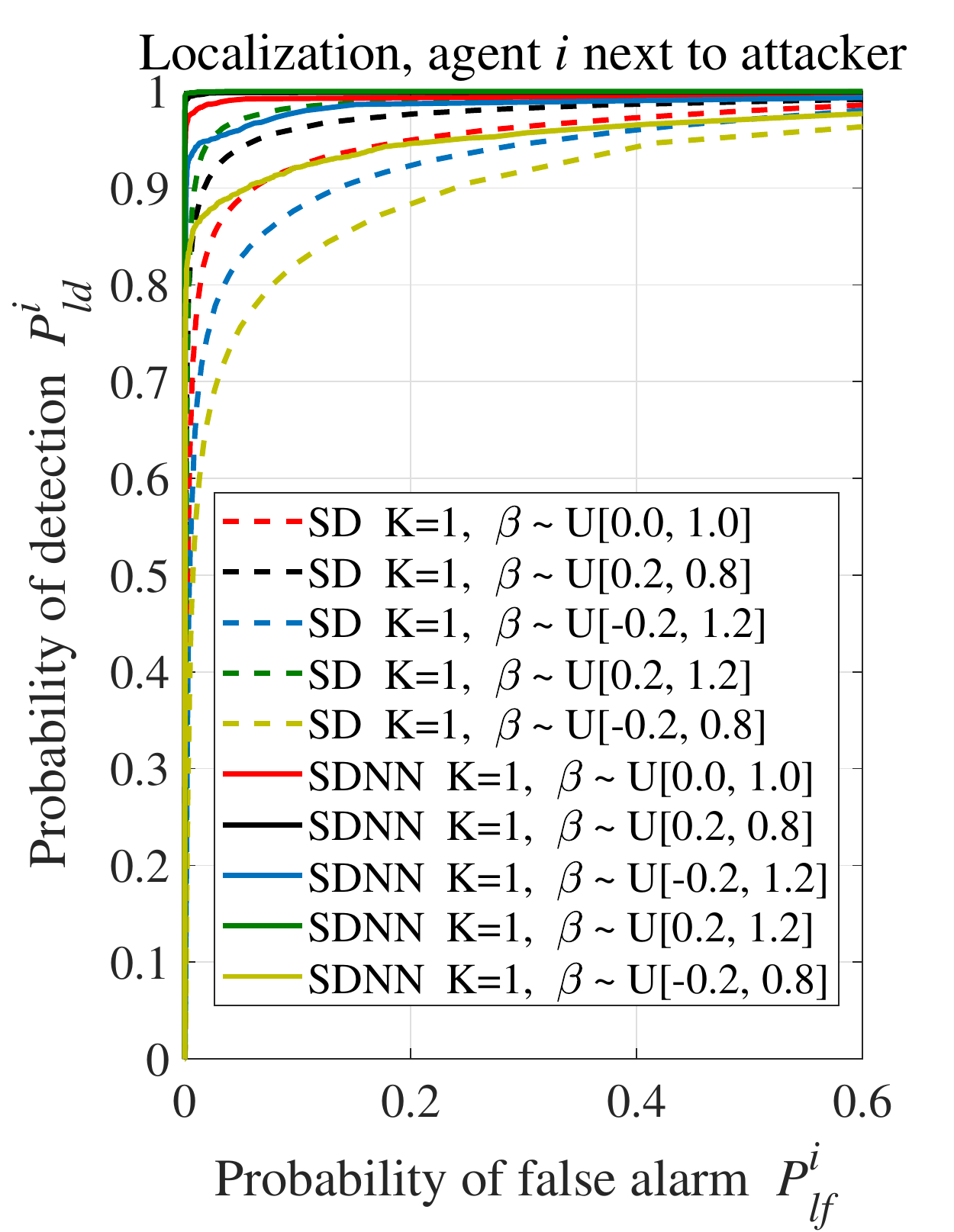}}
	\end{minipage}
	\caption{ROCs of SDNN and SD for the mismatch model: (Left) ND task, (Right) NL task. ${\bm \alpha}^k \sim {\mathcal U}[-0.5,0.5]^d$. Each entry of ${\bm x}^k(0)$ is distributed as legended for testing data.}
	\label{fig:ROC_robust_SDNN1}
\end{figure}
\begin{figure}[t!]
	\begin{minipage}[b]{.49\linewidth}
		\centering
		\centerline{\includegraphics[width=4cm]{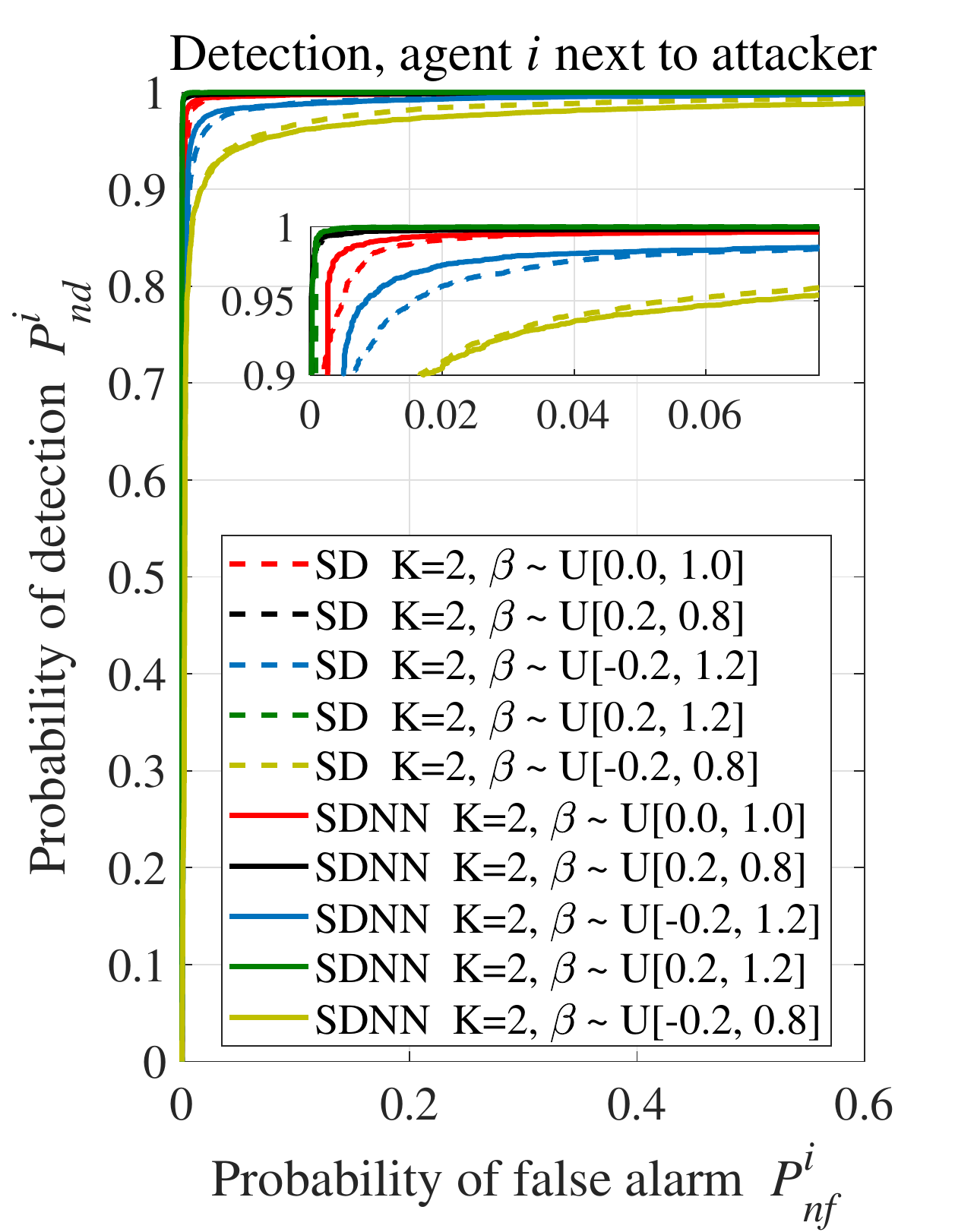}} 
	\end{minipage}
	\hfill
	\begin{minipage}[b]{0.49\linewidth}
		\centering
		\centerline{\includegraphics[width=4cm]{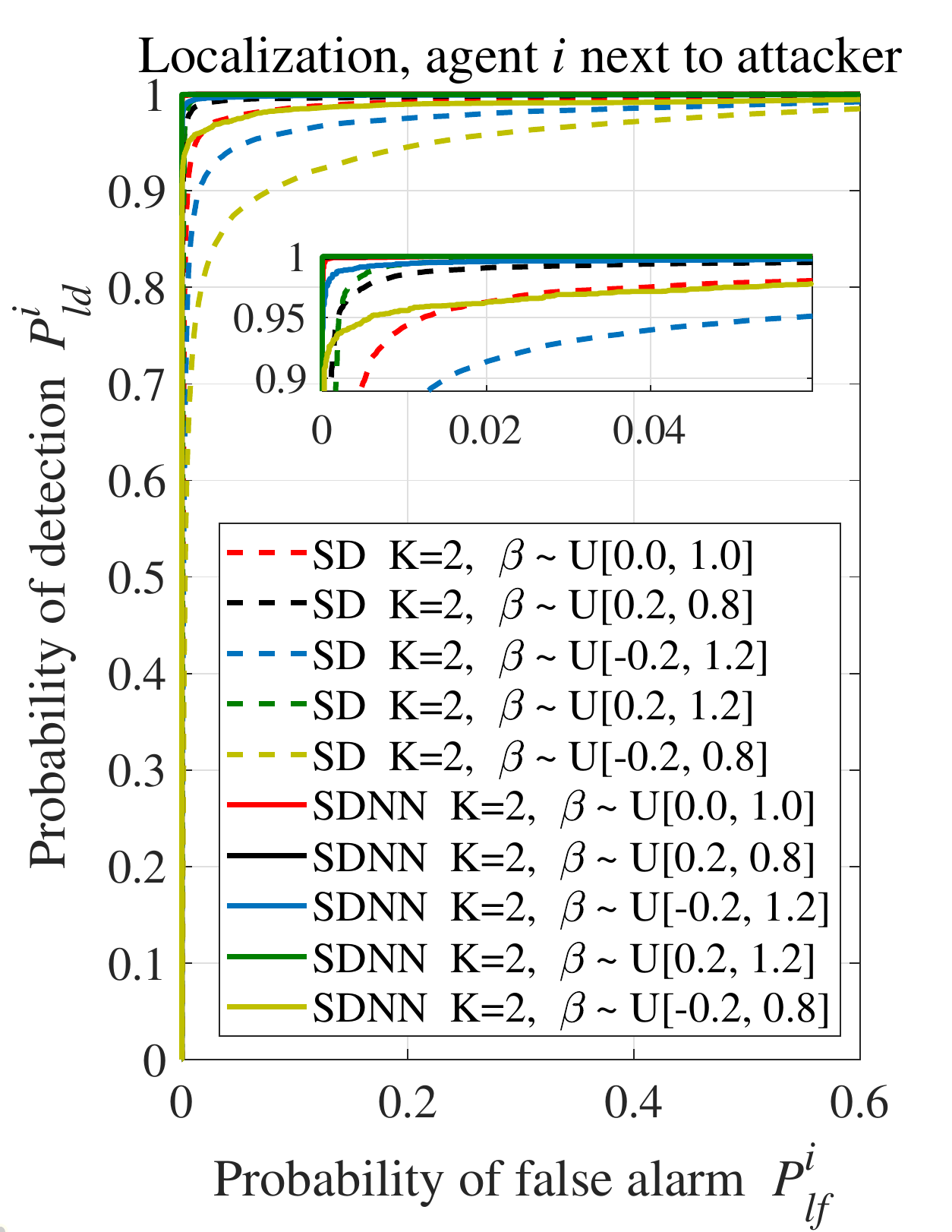}}
	\end{minipage}
	\caption{ROCs of SDNN and SD for the mismatch model: (Left) ND task, (Right) NL task. ${\bm \alpha}^k \sim {\mathcal U}[-0.5,0.5]^d$. Each entry of ${\bm x}^k(0)$ is distributed as legended for testing data.}
	\label{fig:ROC_robust_SDNN2}
\end{figure}
Furthermore, we test the robustness of AI-based methods when the prior information is inconsistent with the actual environment.
The prior information of test scenarios are present in subsection \ref{sec: some_issues}.
We consider one attacker in the Manhattan network and train the parameters of AI-based methods in the scenario with ${\bm \alpha}^k \sim {\mathcal U}[-0.5,0.5]^d$ and $\bm \beta \sim \mc U [0.0, 1.0]^d$, and the test data is generated by other scenarios in TABLE \ref{tab:1}. 

In Fig. \ref{fig:ROC_robust_TDNN}, we show the detection and localization performance of TDNN and TD methods in different test scenarios when $K=5, d=2$. 
Specifically, we generate the test data for the second and third curves by changing the deviation of $\bm \beta$ to $\bm \beta \sim \mc U [0.2, 0.8]^d$ and $\bm \beta \sim \mc U [-0.2, 1.2]^d$.
The results indicate that the performances of TDNN and TD methods deteriorate when the deviation of $\bm \beta$ increases, and improve when the deviation of $\bm \beta$ decreases.
While in the fourth and fifth curves, we change the mean of $\bm \beta$ to $\bm \beta \sim \mc U [0.2, 1.2]^d$ and $\bm \beta \sim \mc U [-0.2, 0.8]^d$.
As can be seen that the performances of TDNN and TD will improve when $[\mathbb{E}[\bm \alpha]- \mathbb{E}[\bm \beta]]$ increases and deteriorate when the gap decreases.
Meanwhile, TDNN performs better than TD in both ND and NL tasks.
In addition, Fig. \ref{fig:ROC_robust_SDNN1} and Fig. \ref{fig:ROC_robust_SDNN2}  respectively show the detection and localization performance of SDNN and SD methods when $K=1, d=2$ and $K=2, d=2$.
In these plots, the ROC curves of SDNN and SD follow the same trends as those in Fig. \ref{fig:ROC_robust_TDNN}.
It is worth mentioning that SDNN still shows good detection and localization performance despite the mismatching of the training data and testing data.
Specifically, in Fig. \ref{fig:ROC_robust_TDNN}, Fig. \ref{fig:ROC_robust_SDNN1} and Fig. \ref{fig:ROC_robust_SDNN2}.

\subsection{Detection and Localization for Multiple Attackers}
\begin{figure}[t!]
\begin{minipage}[b]{.49\linewidth}
  \centering
  \centerline{\includegraphics[width=4cm]{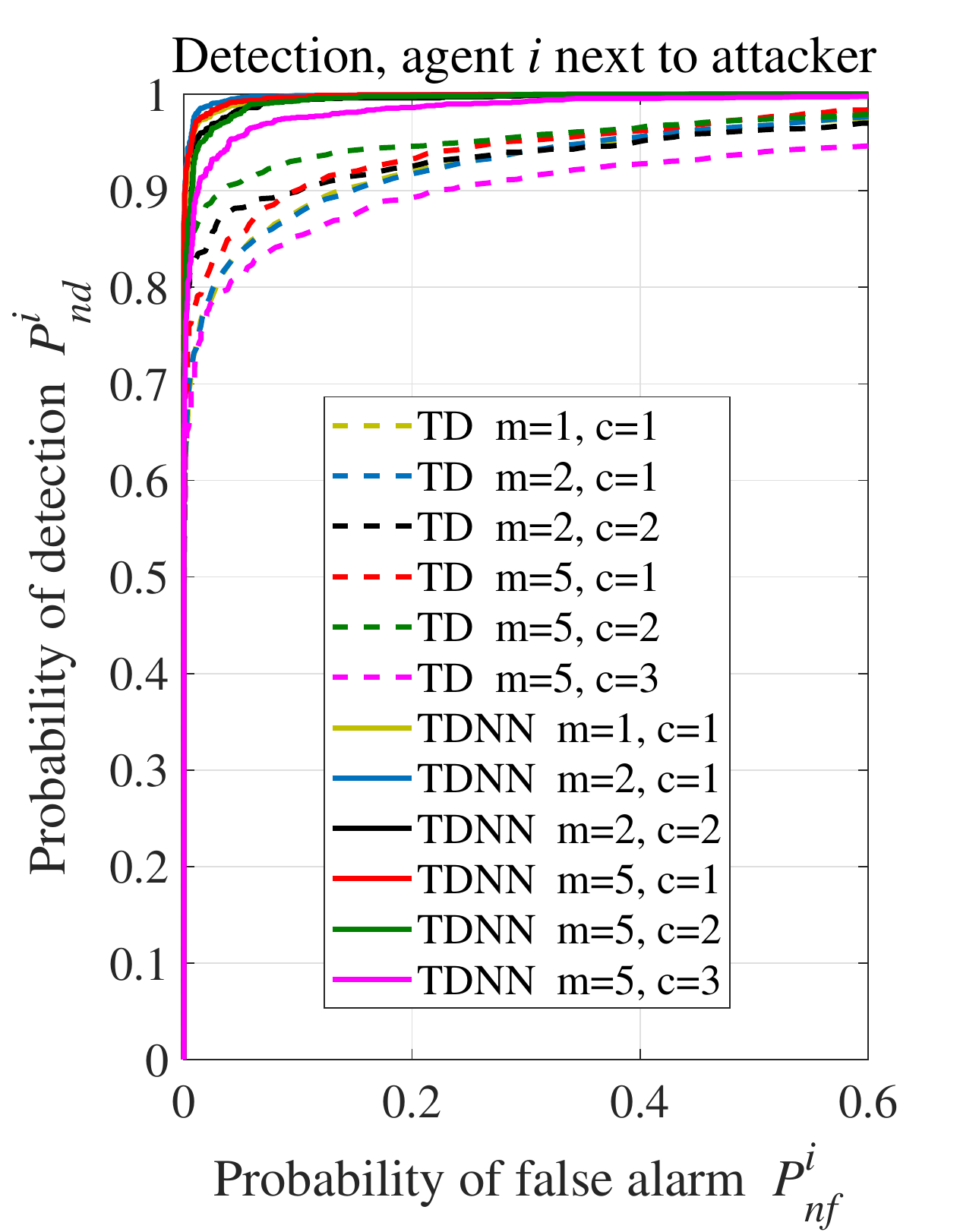}} 
\end{minipage}
\hfill
\begin{minipage}[b]{0.49\linewidth}
  \centering
  \centerline{\includegraphics[width=4cm]{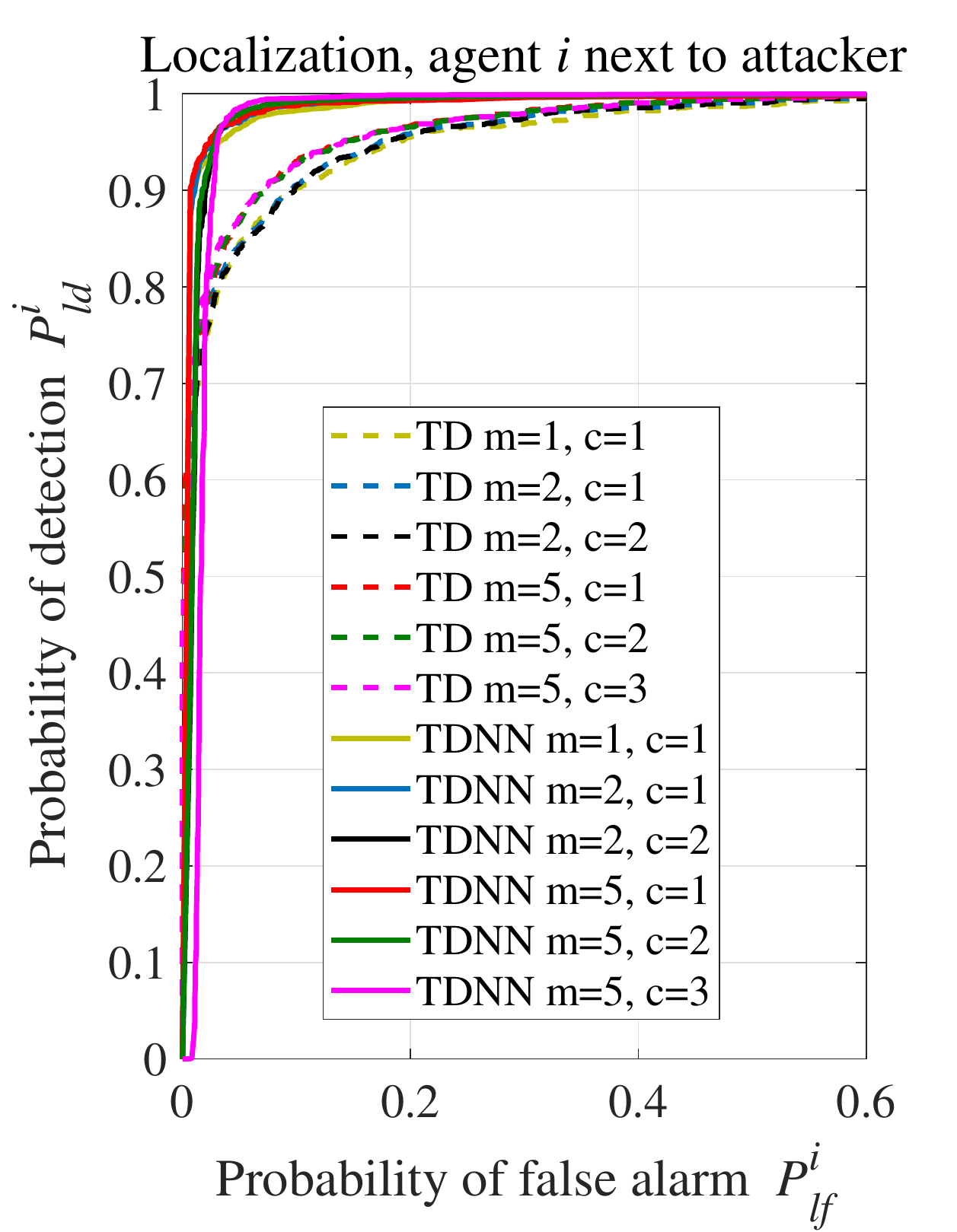}}
\end{minipage}
\caption{ROCs for multiple attackers of TDNN and TD: (Left) ND task, (Right) NL task. $m$ is the number of attackers in the Manhattan network, and $c$ is the number of attackers in the testing agent's neighborhood.}
\label{fig:ROC_multi_TDNN}
\end{figure}
\begin{figure}[t!]
\begin{minipage}[b]{.49\linewidth}
  \centering
  \centerline{\includegraphics[width=4cm]{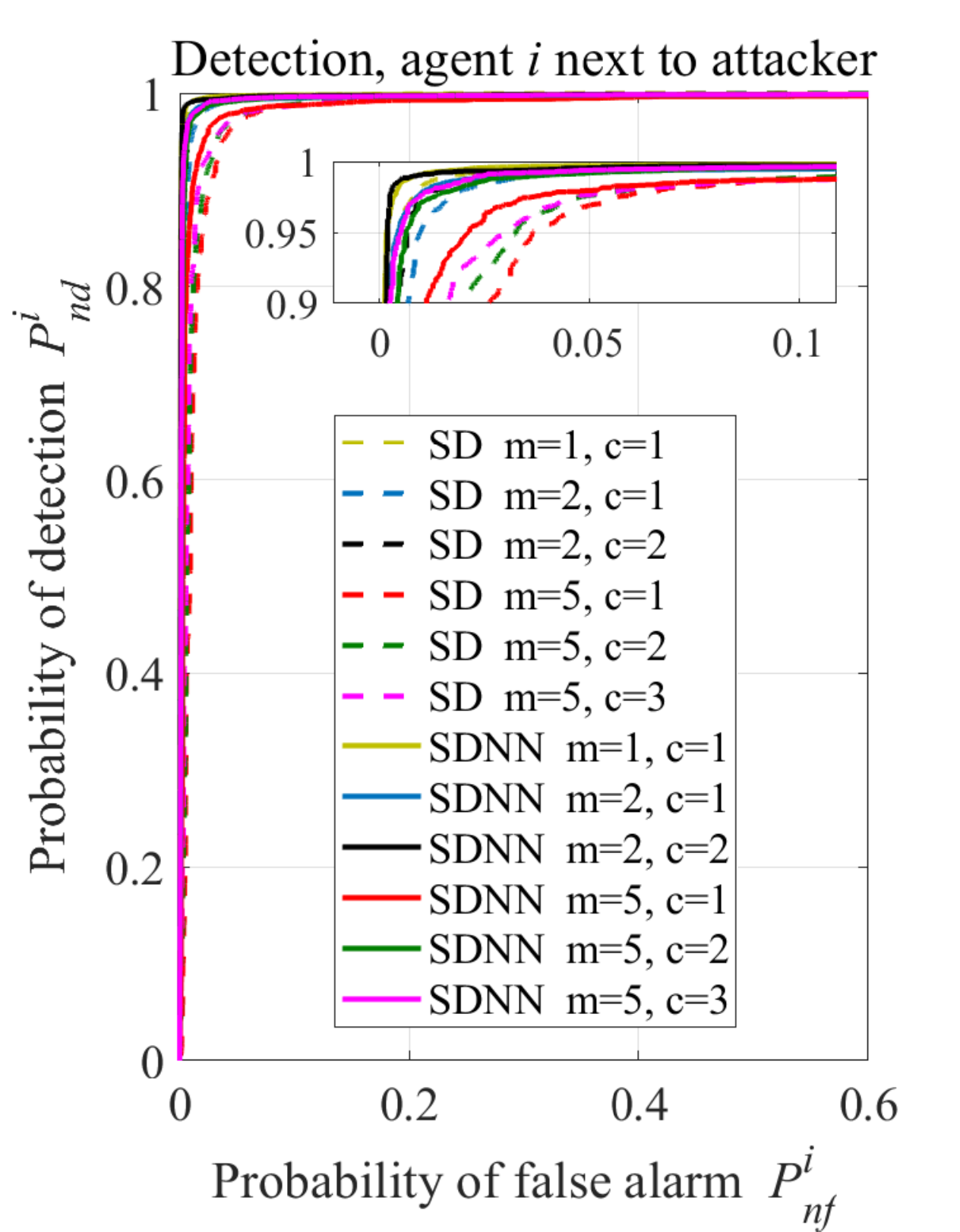}} 
\end{minipage}
\hfill
\begin{minipage}[b]{0.49\linewidth}
  \centering
  \centerline{\includegraphics[width=4cm]{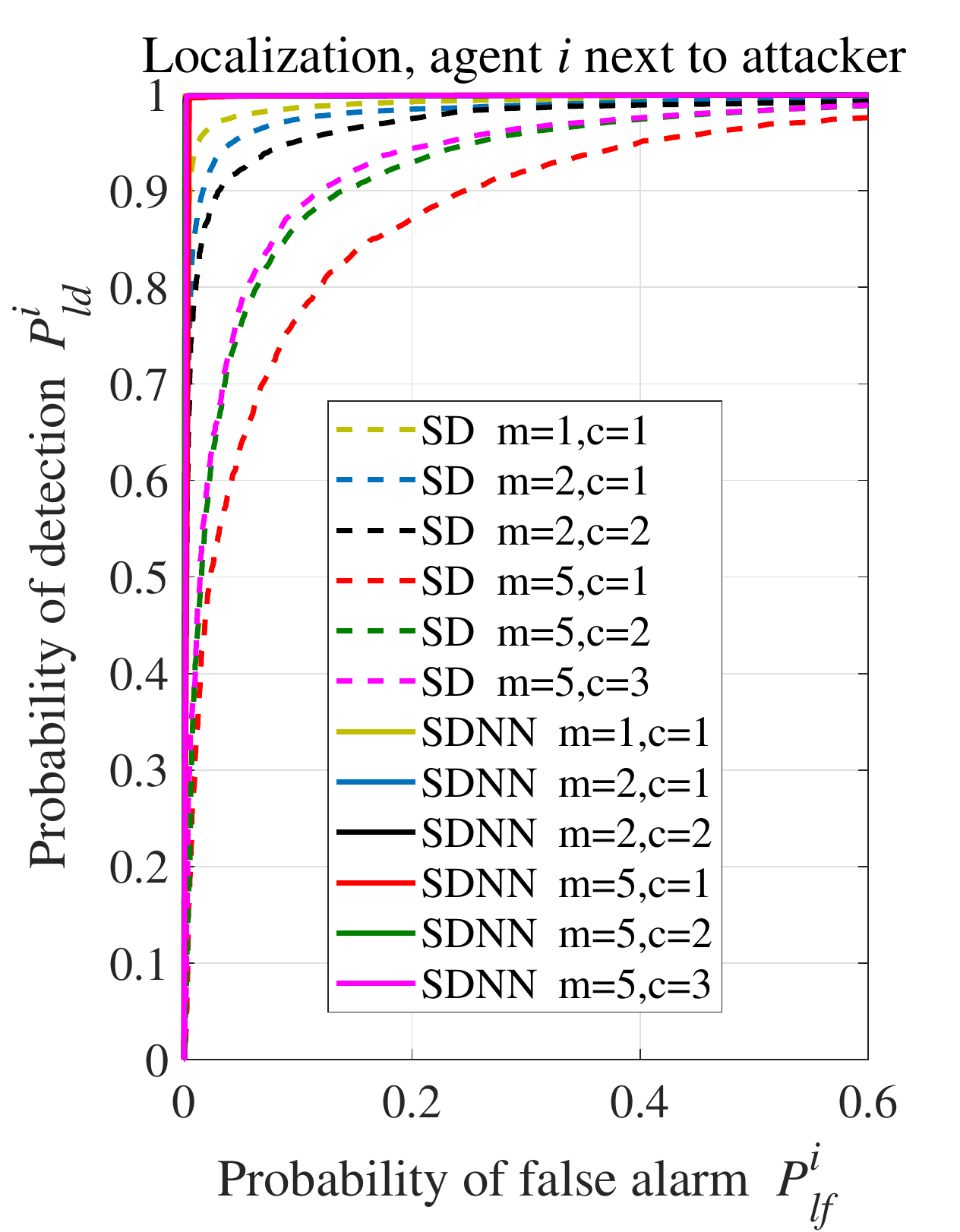}}
\end{minipage}
\caption{ROCs for multiple attackers of SDNN and SD: (Left) ND task, (Right) NL task. $m$ is the number of attackers in the Manhattan network, and $c$ is the number of attackers in the testing agent's neighborhood.}
\label{fig:ROC_multi_SDNN}
\end{figure}
Then, we investigated the performance of TDNN and SDNN with the case of multiple attackers. Note that the parameters of AI-based methods are the same as those in subsection \ref{sec: one_attacker}. 
In Fig. \ref{fig:ROC_multi_TDNN} and \ref{fig:ROC_multi_SDNN}, we set agents $\{1,\cdots,m\}$ as the attackers when considering a scenario with $m$ attackers in the Manhattan network.
The legend in these plots with \emph{`$m$ and $c$'} indicates that there are $m$ attackers in the Manhattan network, and $c$ attackers are in the neighborhood of the monitoring agent. 
In this case, the same ${\bm \alpha}^k$ is shared by all cooperating attackers, but the noise is random and independent among each attacker.

In Fig. \ref{fig:ROC_multi_TDNN}, we show the ROC curves of TDNN and TD methods when $K=5,d=2$. 
Obviously, both the detection and localization performance of TDNN and TD methods fluctuate obviously in different $m$ and $c$.
We notice that the total number of attackers ($m$) has only a slight impact on the detection performance of TDNN, which can be seen from the sixth ($m=1, c=1$), seventh ($m=2, c=1$) and ninth curves ($m=5, c=1$). It shows that the detection performance for TDNN depends on the number of attacking neighbors.
As for NL task, we observe that TDNN exhibits similar performance in different attack scenarios.
Nevertheless, the proposed TDNN method outperforms TD and has good performance in the case of multiple attackers.
For SDNN and SD methods, the detection and localization performance are shown in Fig. \ref{fig:ROC_multi_SDNN} with $K=2, d=2$.
From the Fig. \ref{fig:ROC_multi_SDNN} (Left), SDNN and SD still show good detection performance in the case of multiple attackers. 
We notice that the detection performance of SDNN is a slightly better than SD when $m=5, c=3$.
While in the NL task, SDNN has excellent localization performance and outperforms SD in all attack scenarios.

\subsection{Performance Test In a Small World Network}
\begin{figure}[t!]
\begin{minipage}[b]{.49\linewidth}
  \centering
  \centerline{\includegraphics[width=4cm]{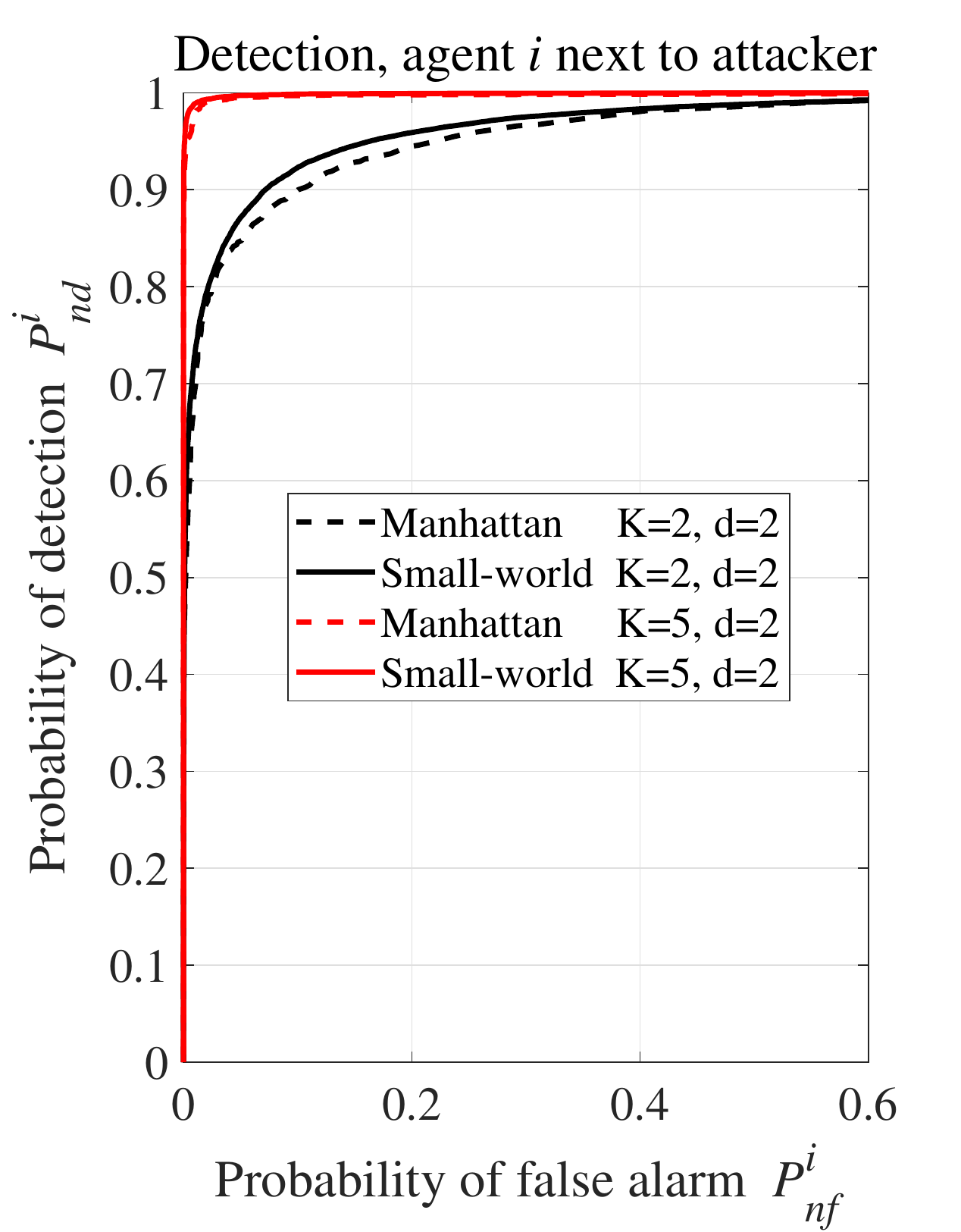}} 
\end{minipage}
\hfill
\begin{minipage}[b]{0.49\linewidth}
  \centering
  \centerline{\includegraphics[width=4cm]{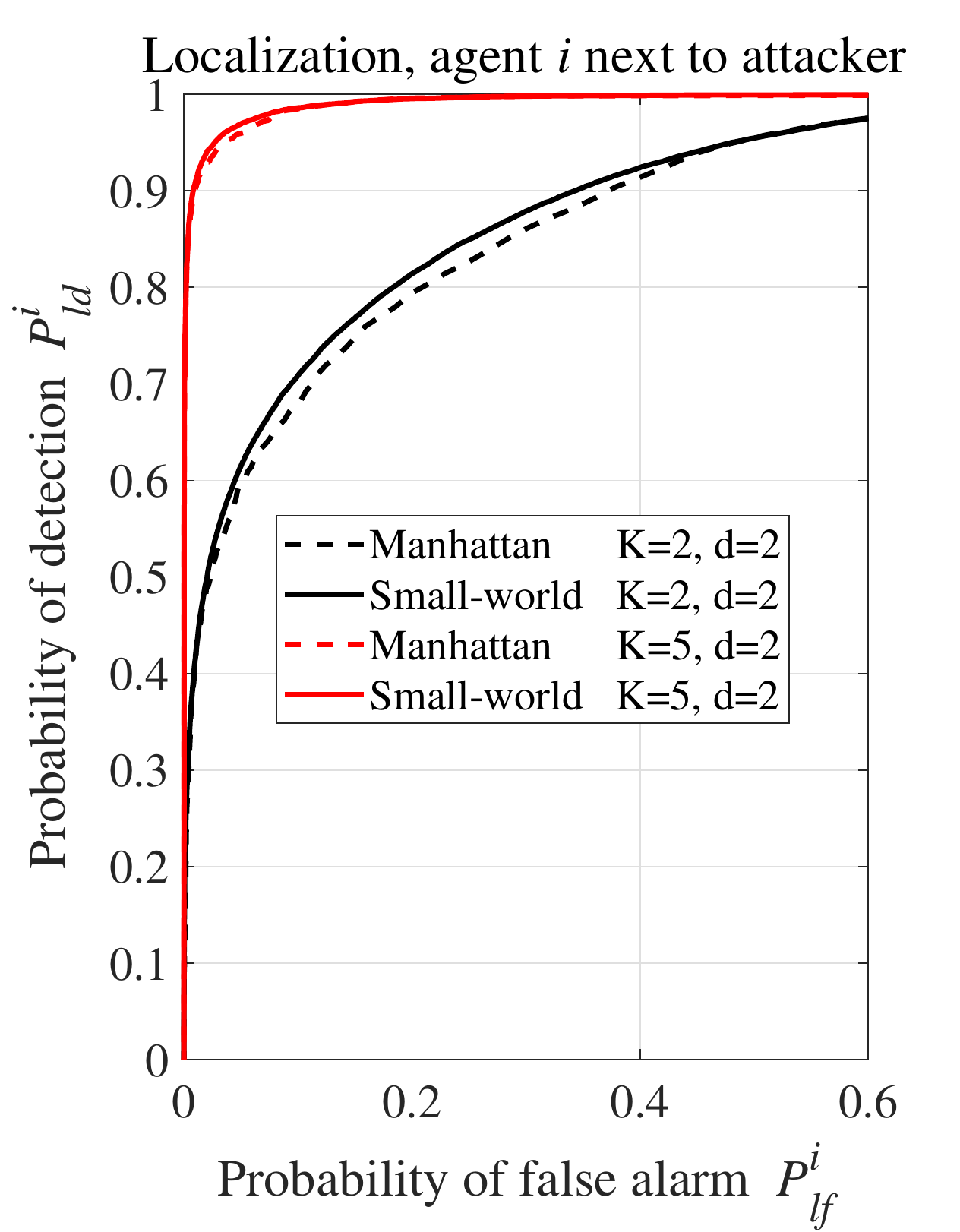}}
\end{minipage}
\caption{ROCs of TDNN for the small world network: (Left) ND task, (Right) NL task. Solid lines show the average detection and localization performance in the small world network, and the parameters of TDNN are trained by the Manhattan network.}
\label{fig:SmallWorld1}
\end{figure}

\begin{figure}[t!]
\begin{minipage}[b]{.49\linewidth}
  \centering
  \centerline{\includegraphics[width=4cm]{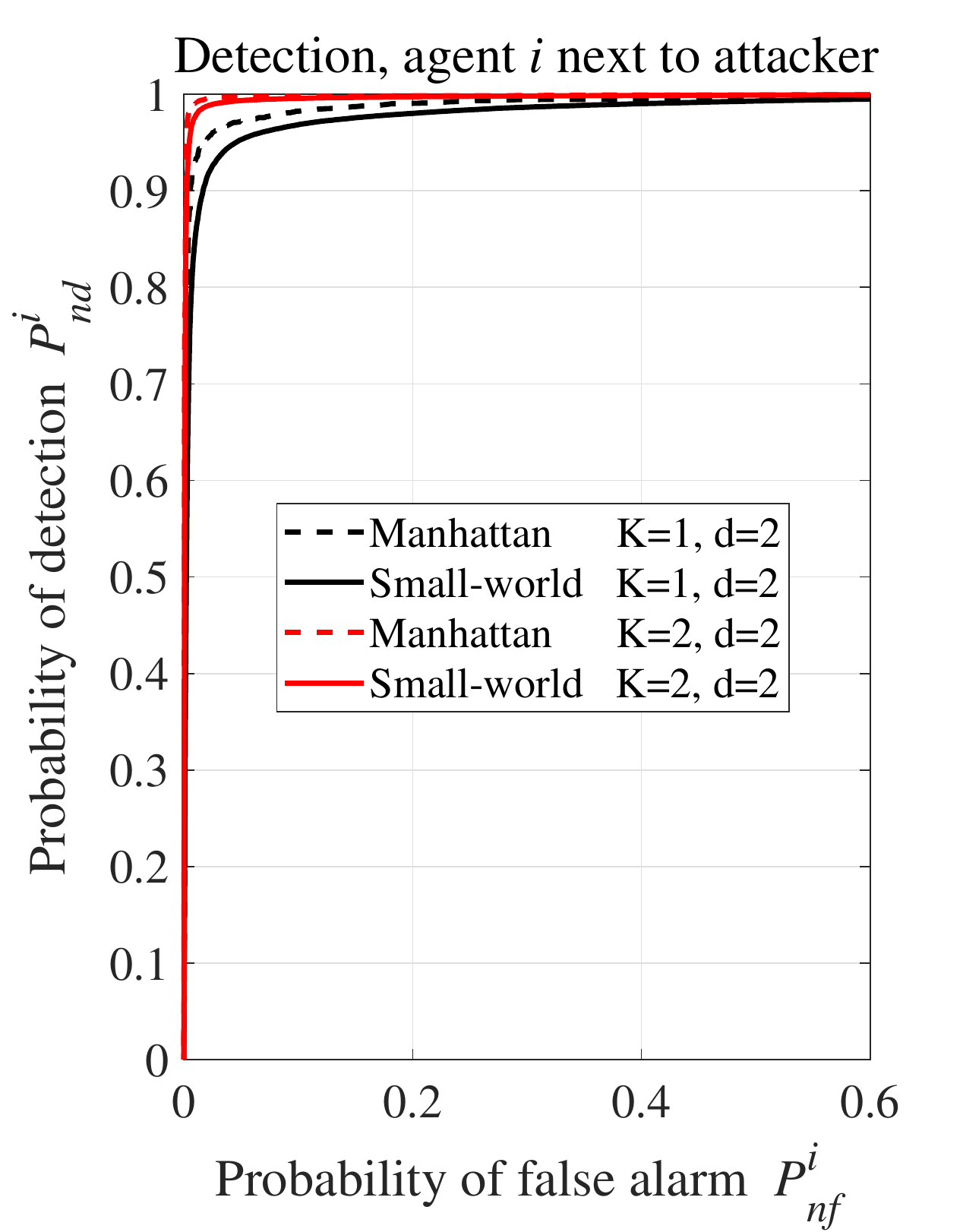}} 
\end{minipage}
\hfill
\begin{minipage}[b]{0.49\linewidth}
  \centering
  \centerline{\includegraphics[width=4cm]{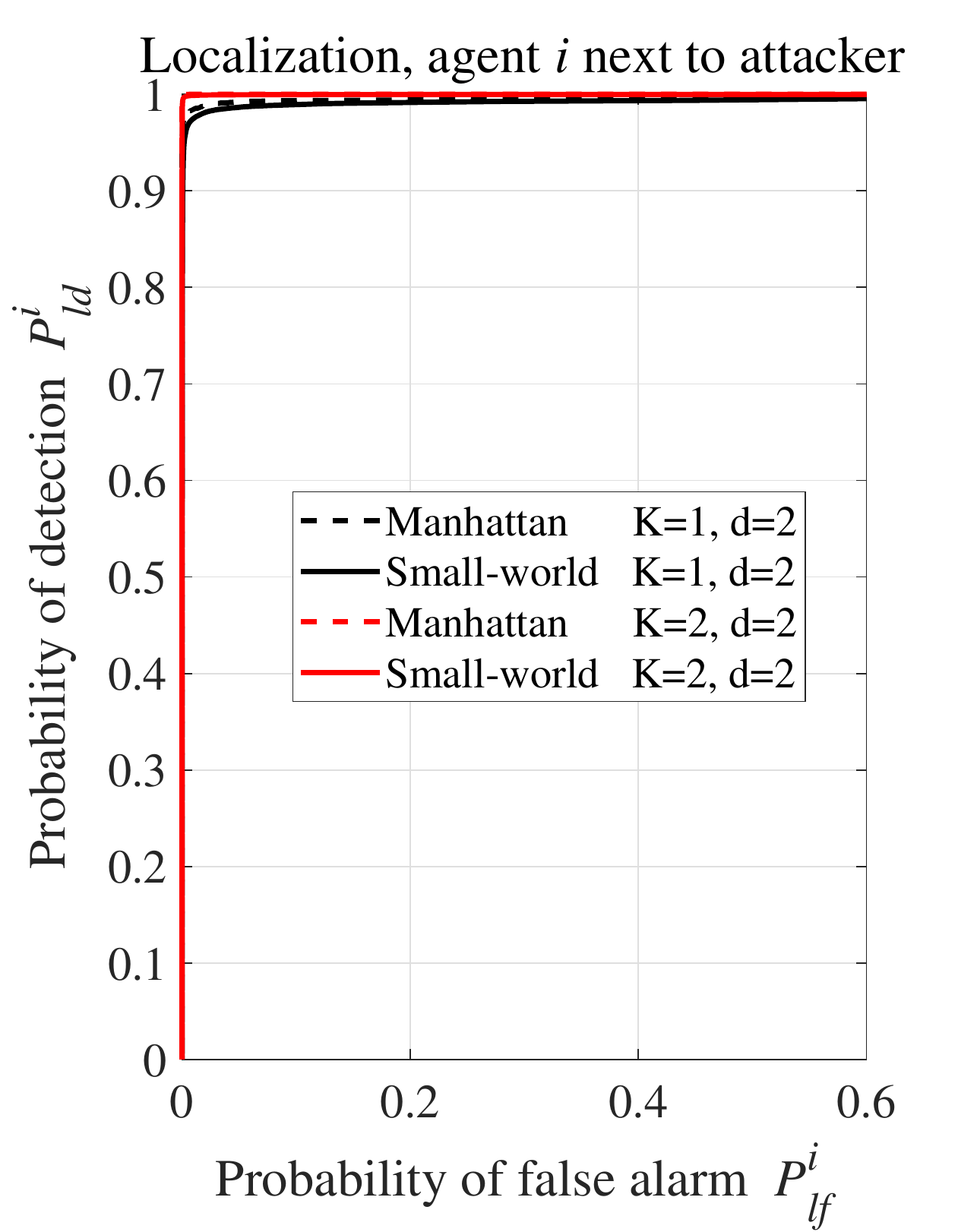}}
\end{minipage}
\caption{ROCs of SDNN for the small world network: (Left) ND task, (Right) NL task. Solid lines show the average detection and localization performance in the small world network, and the parameters of SDNN are trained by the Manhattan network.}
\label{fig:SmallWorld2}
\end{figure}
In addition, we also test the detection and localization performance of AI-based methods in a small world network.
We consider a small world network with $20$ agents. The average degree is set to be $8$ and the rewiring probability is set to be $0.2$. We assume that among all the nodes,  agents $3$, $10$ and $17$ are the attackers. 
The AI-based models are trained by the training data from Manhattan network as those in \ref{sec: one_attacker}, while the test data are collected from the small world network. We only consider to monitor at the agents next to the attacker.
Note that herein the degree-mismatch problem is solved by the proposed method in Section \ref{sec: some_issues}.
We therefore show the performance of TDNN and SDNN in Fig. \ref{fig:SmallWorld1} and Fig. \ref{fig:SmallWorld2} respectively. In these plots, the solid lines show the average detection and localization performance in the small world network.
We notice that AI-based methods also exhibit good detection and localization performance in a small world network.
These results further illustrate the potential of the proposed defense strategies.

\section{Conclusion} \label{sec:conclusion}
This work is dedicated to the detection of insider attacks in the DPS algorithm through AI technology. 
We have proposed two AI-based defense strategies (TDNN and SDNN) for securing the gossip-based DPS algorithm.
Unlike the traditional score-based methods, this work utilizes NN to learn the complex mapping relationships in this classification problem, thus reducing the design difficulty of the attacker detector. To circumvent the mismatch of the training data and the actual network attack, we propose a federated learning approach to learn a local model close to the global model using training data from all agents.
Experiment results demonstrate that the proposed AI-based methods have good detection and localization performance in different attack scenarios. 
They also have good adaptability to different degree of agent, and have strong robustness to the inconsistency of prior information with the actual environment.
Therefore, it is convinced that the proposed AI-based defense strategies have a high potential for practical applications in the DPS algorithm.
As a future work, it would be interesting to try the AI-based methods on more complicated attack models and other decentralized algorithms.


\ifCLASSOPTIONcaptionsoff
  \newpage
\fi
\bibliographystyle{IEEEtran}
\bibliography{reference}

\end{document}